\documentclass[12pt]{article}

\usepackage{newtxtext,newtxmath}
\usepackage{algorithm}
\usepackage{algpseudocode}

\usepackage{amsmath,amssymb}
\usepackage{adjustbox}

\usepackage{graphicx}

\usepackage[letterpaper,margin=1in]{geometry}

\renewenvironment{abstract}
	{\quotation}
	{\endquotation}

\date{}

\makeatletter
\renewcommand{\fnum@figure}{\textbf{Figure \thefigure}}
\renewcommand{\fnum@table}{\textbf{Table \thetable}}
\makeatother

\usepackage{scicite}

\usepackage{url}

\def\scititle{
	Real-time autonomous magnetic microrobot navigation across dynamic and biologically relevant environments
}
\title{\bfseries \boldmath \scititle}

\author{
	Yanda Yang$^{1}$,
	Max Sokolich$^{1}$,
	Fatma Ceren Kirmizitas$^{1,2}$,
    Baylen Ravenscraft$^{1}$,\and
    Sambeeta Das$^{1\ast}$,
    Andreas A. Malikopoulos$^{3}$\and
	\small$^{1}$Department of Mechanical Engineering, University of Delaware, Newark 19716, USA.\and
	\small$^{2}$Department of Animal and Food Sciences, University of Delaware, Newark 19716, USA.\and
    \small$^{3}$School of Civil and Environmental Engineering, Cornell University, Ithaca, NY 14853, USA.\and
	\small$^\ast$Corresponding author. Email: samdas@udel.edu
}

\begin{document} 

\maketitle

\begin{abstract} \bfseries \boldmath
Autonomous microrobots could enable minimally invasive interventions in confined biological environments, but their operation requires real-time navigation among moving obstacles and environmental disturbances. Here we present a closed-loop framework for autonomous magnetic microrobot navigation that separates long-range geometric planning from short-range reactive control. An analytic geometry-based planner continuously generates collision-free global routes from microscopy images, while rule-based or reinforcement-learning local controllers respond to moving obstacles before returning control to the replanned global route. The planner produces short paths with substantially lower computation time than representative graph-, sampling- and geometry-based planners, and the integrated framework operates within the video-rate control budget. We demonstrate microrobot autonomous navigation in static and dynamic microfluidic environments, including avoidance of moving particles and human red blood cells, navigation with and against fluid flow, and repeated long-duration target transitions. We further demonstrate magnetic microrobot actuation in ex vivo ocular and vascularized cortical tissues and photoacoustic detection of magnetically induced microrobot dynamics in post-mortem tissue. These results establish a modular framework for autonomous magnetic microrobot navigation in complex biological environments.
\end{abstract}

\noindent

Microrobots offer a route to minimally invasive manipulation and intervention in confined biological environments that are difficult to access with conventional instruments \cite{nelson2010microrobots,ceylan2019translational,yang2026micropush,yang2026residual}. Magnetic microrobots are particularly attractive because magnetic fields can transmit force and torque remotely through biological materials without requiring onboard power, enabling locomotion, cargo transport and cellular manipulation at the microscale \cite{du2023image,yang2020magnetic,yang2023rolling,yang2024quadrupole}. Yet remote actuation alone does not enable autonomous microrobot navigation. The small dimensions that provide access to confined environments also limit conventional onboard sensing, computation and actuation \cite{chowdhury2015controlling,jiang2022control}. Navigation must therefore rely on an external closed loop in which the environment is observed, a route is generated, motion commands are applied and the resulting robot state is continuously fed back to guide subsequent actions \cite{hou2023review,yang2023closed}.

Establishing such a loop becomes difficult in biological environments. A microrobot must make long-range progress through boundaries, branches and narrow passages while maintaining clearance from nearby structures, but the geometry observed at one instant may not remain valid during execution \cite{landers2025clinically}. Cells, particles and other suspended objects can move with local flow, and disturbances in sensing and actuation become significant relative to the dimensions of the robot \cite{jia2021robust}. Autonomous navigation therefore poses coupled problems at different spatial and temporal scales: global planning must identify an efficient route through the accessible workspace, whereas local control must respond rapidly when the immediate environment changes \cite{li2017autonomous,yang2021autonomous}. Treating either problem in isolation can produce a geometrically valid route that becomes unsafe during execution, or a reactive trajectory that avoids nearby obstacles but fails to make consistent progress toward the target \cite{wang2024deep,zhong2026adaptive,yang2022hierarchical}.

Existing microrobot path-planning approaches include graph-search, sampling-based and optimization-based \cite{yang2019automated,liu2019navigation,salehizadeh2021path,wang2021micromanipulation}. Grid-based methods such as A* provide systematic search but their computational and memory costs increase as spatial resolution is refined \cite{warren1993fast}. Sampling-based planners such as rapidly exploring random trees can operate directly in continuous spaces but introduce stochastic variation in computation time and path quality, particularly in constrained passages \cite{karaman2011sampling}. Optimization-based approaches can balance competing planning objectives but often require repeated evaluations that limit rapid replanning \cite{yang2022autonomous}. These trade-offs are especially consequential for microscopy-guided microrobots, for which planning must be repeated at rates comparable to image acquisition rather than performed once before execution \cite{zou2022dynamic}.

Dynamic obstacle avoidance introduces a complementary challenge. Reactive approaches including artificial potential fields, velocity-based methods, the dynamic window approach and learning-based controllers can modify motion in response to nearby moving objects \cite{fan2022obstacle,kim2017autonomous,jiang2023dqn,yang2022autonomous}. However, local avoidance alone does not determine how the microrobot should progress through the larger environment. Conversely, repeatedly replanning a collision-free geometric path from successive images does not explicitly account for motion that occurs between planning and control updates. End-to-end learned navigation could in principle couple these decisions, but it places global routing, local adaptation and constraint handling within a single policy and requires training data that adequately represent the range of geometries and disturbances encountered during operation. A modular global–local architecture offers an alternative: deterministic planning can preserve interpretable long-range progress, while learning or reactive control can be restricted to the short-horizon interactions for which adaptation is most useful \cite{jian2021global,wang2020dynamics}.

A further gap separates algorithmic navigation studies from operation in biological tissue. Many microrobot navigation systems are developed under direct optical observation in simplified chambers or microfluidic channels. These platforms are essential for controlled evaluation but do not reproduce the optical accessibility, heterogeneous structure and sensing constraints expected in tissue. Progress toward biological autonomy therefore requires not only increasingly robust planning and control, but also evidence that magnetic actuation remains effective in complex biological media and that microrobots can ultimately be detected when direct microscopy is unavailable. Photoacoustic imaging is promising in this respect because it combines optical contrast with ultrasonic detection at depths beyond conventional optical microscopy \cite{landers2025clinically,li2018development,jeon2019magnetically,han2024imaging,aziz2020medical,xie2020photoacoustic}.

Here we develop a real-time closed-loop framework for autonomous magnetic microrobot navigation that separates global geometric reasoning from local response to environmental change. An analytic geometry-based planner (AGP) repeatedly constructs a global collision-free route from the current image, while either an interpretable rule-based controller or a reinforcement-learning controller temporarily modifies the motion during short-range encounters with moving obstacles. We benchmark the global planner against representative graph-, sampling- and geometry-based approaches and evaluate the global–local architecture across increasingly complex dynamic environments. We then integrate perception, planning and control with a three-axis magnetic actuation system and demonstrate autonomous rolling-microrobot navigation around moving particles and human red blood cells, with and against imposed flow, and during repeated long-duration target transitions. Finally, we extend the experimental scope beyond transparent microfluidic environments by demonstrating magnetic microrobot motion in ex vivo mouse eye and vascularized cortical tissue and photoacoustic detection of magnetically induced microrobot dynamics in post-mortem tissue. Together, these results connect real-time navigation algorithms with physical actuation and emerging biological sensing modalities toward autonomous microrobot navigation in complex environments.

\section*{RESULTS}

\subsection*{Autonomous microrobot navigation framework}

\begin{figure}
    \centering
    \includegraphics[width=0.75\textwidth]{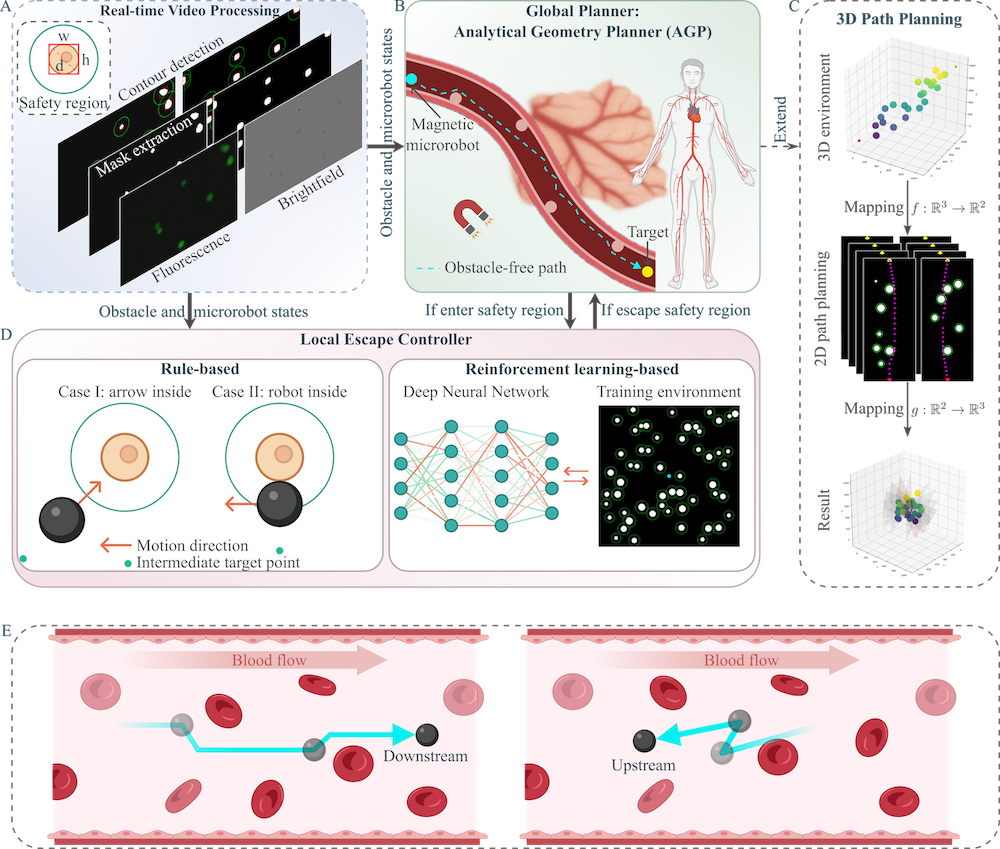}
    \caption{\textbf{Closed-loop framework for autonomous magnetic microrobot navigation.} (\textbf{A}) Real-time microscopy and image processing extract the microrobot and obstacle states and construct configuration-space safety regions. (\textbf{B}) The analytic geometry-based planner (AGP) provides a global route to the target. (\textbf{C}) Overview of the three-dimensional extension based on planar projection, per-slice planning, and reconstruction; the full formulation is provided in Supplementary Methods. (\textbf{D}) Reactive local escape is activated near moving obstacles using either an interpretable rule-based controller or a reinforcement-learning-based controller, after which control returns to the global planner. (\textbf{E}) Representative operating concept for downstream and upstream navigation in a vascular-like environment containing moving cells.}
    \label{fig:framework}
\end{figure}

We developed a closed-loop magnetic microrobot navigation framework that combines real-time microscopy, deterministic geometric planning, reactive local obstacle avoidance, and magnetic actuation (Figure~\ref{fig:framework}). The framework is organized around a global--local decomposition. Image processing converts each incoming frame into the microrobot state, obstacle geometry, and target location; the analytic geometry-based planner (AGP) then computes a collision-free route from the current state. When nearby obstacle motion creates a short-horizon conflict, a local escape controller temporarily overrides the nominal route. After the conflict is resolved, control returns to AGP and the route is recomputed from the updated image. This decomposition allows long-range progress and short-range reactivity to be handled by different modules while preserving a transparent geometric global plan.

The same architecture is used across static, dynamic, and flowing environments. In the benchtop system, camera feedback closes the loop between perception and a three-axis Helmholtz coil system that actuates a rolling magnetic microrobot. The planner and local controller are updated repeatedly from the live image stream rather than executing a precomputed open-loop trajectory. Figure~\ref{fig:framework}(E) illustrates the intended operating regime in a vascular-like environment, where the microrobot must continue making progress to target while cells and other suspended objects move with the flow. A three-dimensional extension of AGP based on planar slicing and path reconstruction is summarized in Figure~\ref{fig:framework}(C), with the full formulation provided in Supplementary Methods and Supplementary Fig.~\ref{fig:3D}; the main text focuses on the experimentally validated two-dimensional closed-loop system.

\subsection*{Analytic geometry-based planner}

\begin{figure}
    \centering
    \includegraphics[width=0.75\textwidth]{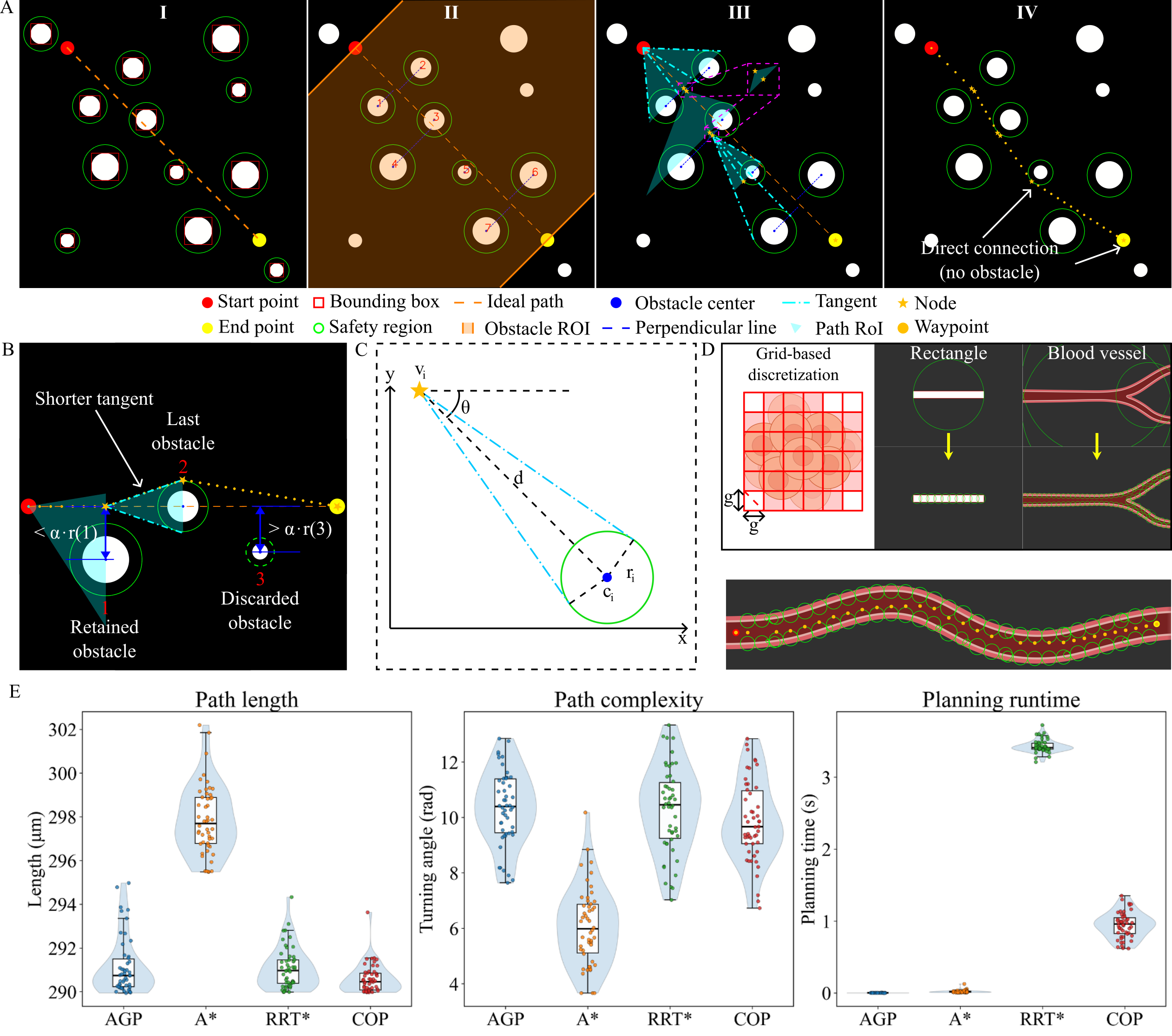}
    \caption{\textbf{Analytic geometry-based planning and static-planner benchmark.} (\textbf{A}) Sequential AGP construction from the direct start--goal line to a collision-free route through inflated obstacles. (\textbf{B}) Obstacle pruning using the distance factor $\alpha$, showing retained and discarded obstacles. (\textbf{C}) Tangent construction used to advance the geometric route around an active safety region. (\textbf{D}) Representation of noncircular structures and vessel boundaries by multiple local safety regions. (\textbf{E}) Comparison of AGP with weighted A*, RRT*, and circular-obstacle pathfinding (COP) across the 50 paired benchmark environments; all four planners returned valid paths in the full test set. Violin plots show the distributions of path length, total absolute turning angle, and wall-clock planning time, with scene-level results and embedded boxplots.}
    \label{fig:AGP}
\end{figure}

AGP converts the detected microrobot and obstacle geometries into a geometric path-planning problem by inflating each obstacle according to the microrobot radius, allowing the microrobot to be treated as a point moving among safety regions (Figure~\ref{fig:AGP}(A)). Starting from the direct start--goal line, obstacles unlikely to constrain the route are first removed using their projected positions and distances from the ideal path (Figure~\ref{fig:AGP}(B)).

For the remaining obstacles, AGP constructs a sparse sequence of primary nodes using perpendicular projections and tangent lines around nearby safety regions (Figure~\ref{fig:AGP}(A,C)). The construction is advanced sequentially toward the target, selecting geometrically valid tangent or projection-based nodes that remain outside the active safety regions. Once the sparse route is obtained, additional waypoints are inserted along the path for closed-loop tracking.

Extended or noncircular structures are represented by multiple local safety regions rather than a single large enclosing region, preserving navigable free space near curved boundaries while allowing the same geometric construction to be applied (Figure~\ref{fig:AGP}(D)). Full mathematical definitions and implementation details are provided in Methods.

\subsection*{Benchmarking against established planners}

We next asked whether the low computational cost of AGP was obtained at the expense of trajectory quality. AGP was compared with weighted A*, RRT*, and circular-obstacle pathfinding (COP), representing raster search, sampling-based planning, and geometry-aware graph search, respectively (Figure~\ref{fig:AGP}(E)). All four planners returned valid collision-free paths in all evaluated environments.

AGP produced short paths with an average length of approximately $291\,\mu\mathrm{m}$, comparable to RRT* and COP and shorter than weighted A* in the evaluated implementation. Weighted A* yielded lower cumulative turning, whereas AGP, RRT*, and COP showed comparable turning-angle distributions associated with continuous geometric detours around circular obstacles. The clearest separation was computational: AGP required approximately $1.25\,\mathrm{ms}$ per planning query, compared with tens of milliseconds for weighted A*, approximately one second for COP, and several seconds for RRT*. Thus, the goal is not global path optimality, but a practical combination of deterministic construction, competitive path quality, and millisecond-scale replanning compatible with closed-loop imaging.

\subsection*{Reactive local control for moving obstacles}

\begin{figure}
    \centering
    \includegraphics[width=0.75\textwidth]{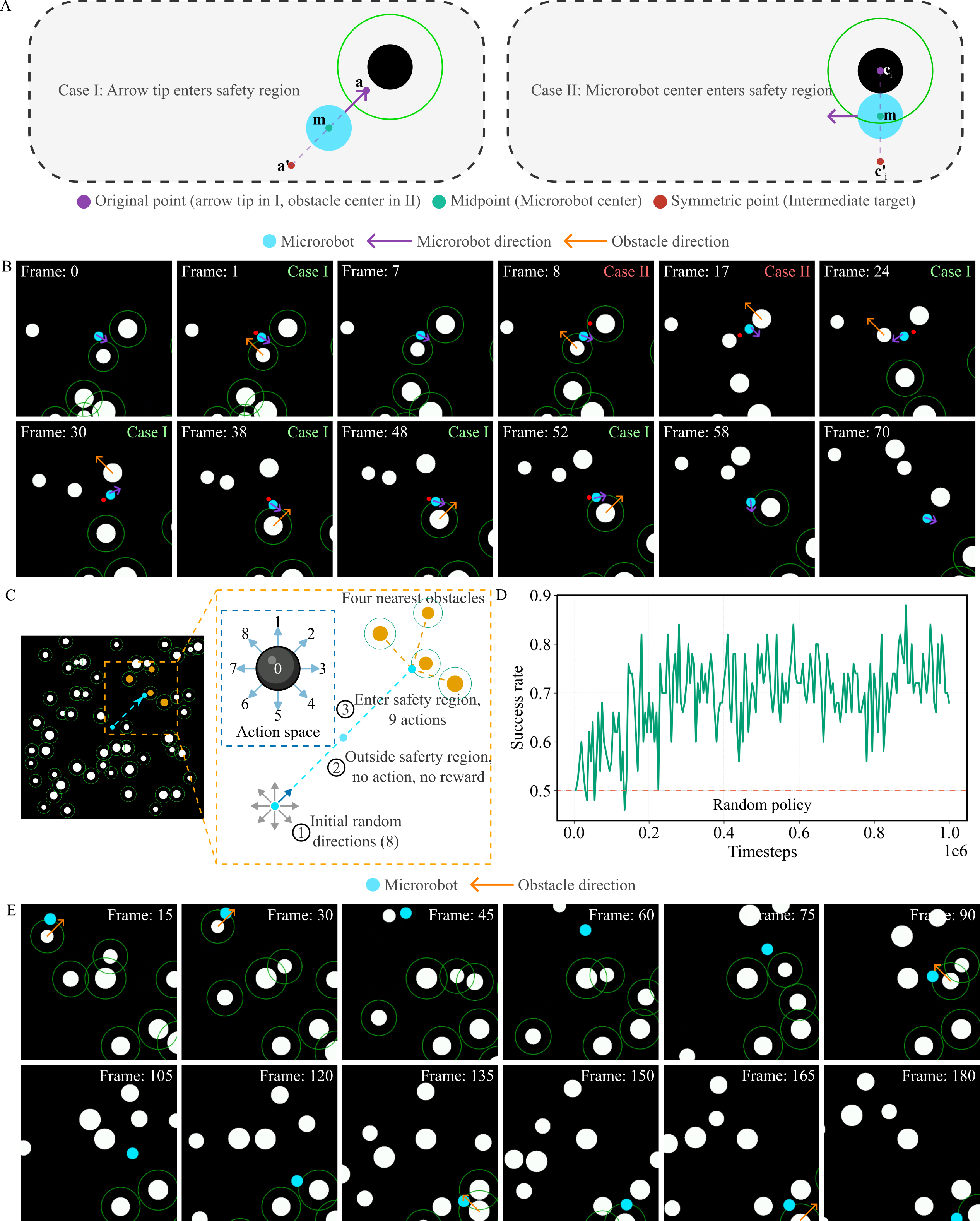}
    \caption{\textbf{Reactive local escape controllers for dynamic obstacles. See Supplementary Video~4.} (\textbf{A}) Geometric rule-based controller. Case~I reflects the predicted motion point when it enters a safety region; Case~II reflects the obstacle center when the microrobot itself enters the safety region. (\textbf{B}) Representative rule-based avoidance sequence showing repeated local interventions and subsequent escape. (\textbf{C}) Reinforcement-learning training environment and nine-action local action space; the policy acts only after entering a safety region. (\textbf{D}) Validation success during DQN training over $10^6$ environment steps; the dashed line indicates the random-policy baseline. (\textbf{E}) Representative learned local-avoidance sequence in a cluttered dynamic scene.}
    \label{fig:local_controller}
\end{figure}

Frame-by-frame replanning updates the global route as the environment changes, but a path that is collision-free in the current image can still become unsafe before the microrobot reaches the next waypoint if a nearby obstacle moves rapidly. We therefore augment AGP with two local controllers that operate only during short-range conflicts (Figure~\ref{fig:local_controller}). The local module does not replace the global planner: its role is to resolve the immediate interaction, after which control returns to the current AGP route.

The first controller uses a deterministic geometric escape rule and requires no training. Depending on whether the predicted one-step microrobot motion or the microrobot itself enters an obstacle safety region, a temporary escape target is constructed by reflecting either the predicted motion point or the obstacle center about the microrobot position (Figure~\ref{fig:local_controller}(A,B)). The temporary target is updated as the relative geometry changes and is released once the local conflict is cleared.

The second controller learns short-range escape actions using deep Q-learning. The learned policy is deliberately restricted to local obstacle interactions and does not observe the global target or AGP path. It acts only when the microrobot enters a local safety region and selects among nine discrete motion actions based on the relative positions and velocities of nearby obstacles (Figure~\ref{fig:local_controller}(C)). This formulation restricts the learning problem to local collision avoidance rather than requiring the policy to learn end-to-end navigation from the current state to a distant target.

Validation success increased above the random-policy baseline during training, with the strongest checkpoints approaching $90\%$ success (Figure~\ref{fig:local_controller}(D)). A representative learned avoidance sequence is shown in Figure~\ref{fig:local_controller}(E). The complete controller definitions, reward formulation, and DQN training procedure are provided in Methods and Supplementary Tables~S1--S3.

\subsection*{Global--local dynamic navigation}

We evaluated the complete global--local architecture in a $2000\times2000$-pixel workspace containing both moving and static circular obstacles (Supplementary Table~S1). The navigation evaluation used the same obstacle counts, obstacle-size range, and microrobot parameters. Moving obstacles traveled at $4$--$8$ pixels per step along one of 16 uniformly spaced directions and reflected from workspace boundaries and other obstacles. AGP used $\alpha=6$ for obstacle pruning in these dynamic scenes.

The evaluated configurations were AGP alone, AGP combined with the rule-based controller, AGP combined with the RL-based controller, and each local controller without AGP. In the coupled configurations, AGP replanned from the latest obstacle state at every simulation step, while the selected local controller temporarily replaced the nominal command when its activation condition was met. The local-controller-only configurations were used to isolate the contribution of global planning.

A benefit of using AGP online rather than as a one-time planner is that the remaining route can be updated continuously from the current microrobot position. In the representative static example in Supplementary Fig.~\ref{fig:AGP_online_static}, successive replanning shortened the executed route relative to following the initial solution as a fixed path. Online replanning also provides a degree of adaptability to dynamic environments. In several representative scenarios involving randomly moving obstacles, coherently translating obstacles, moving targets, and vascular-like geometries, repeated global replanning alone was sufficient to reach the target (Supplementary Fig.~\ref{fig:AGP_online_dynamic}; Supplementary Video~2).

\begin{figure}
    \centering
    \includegraphics[width=0.75\textwidth]{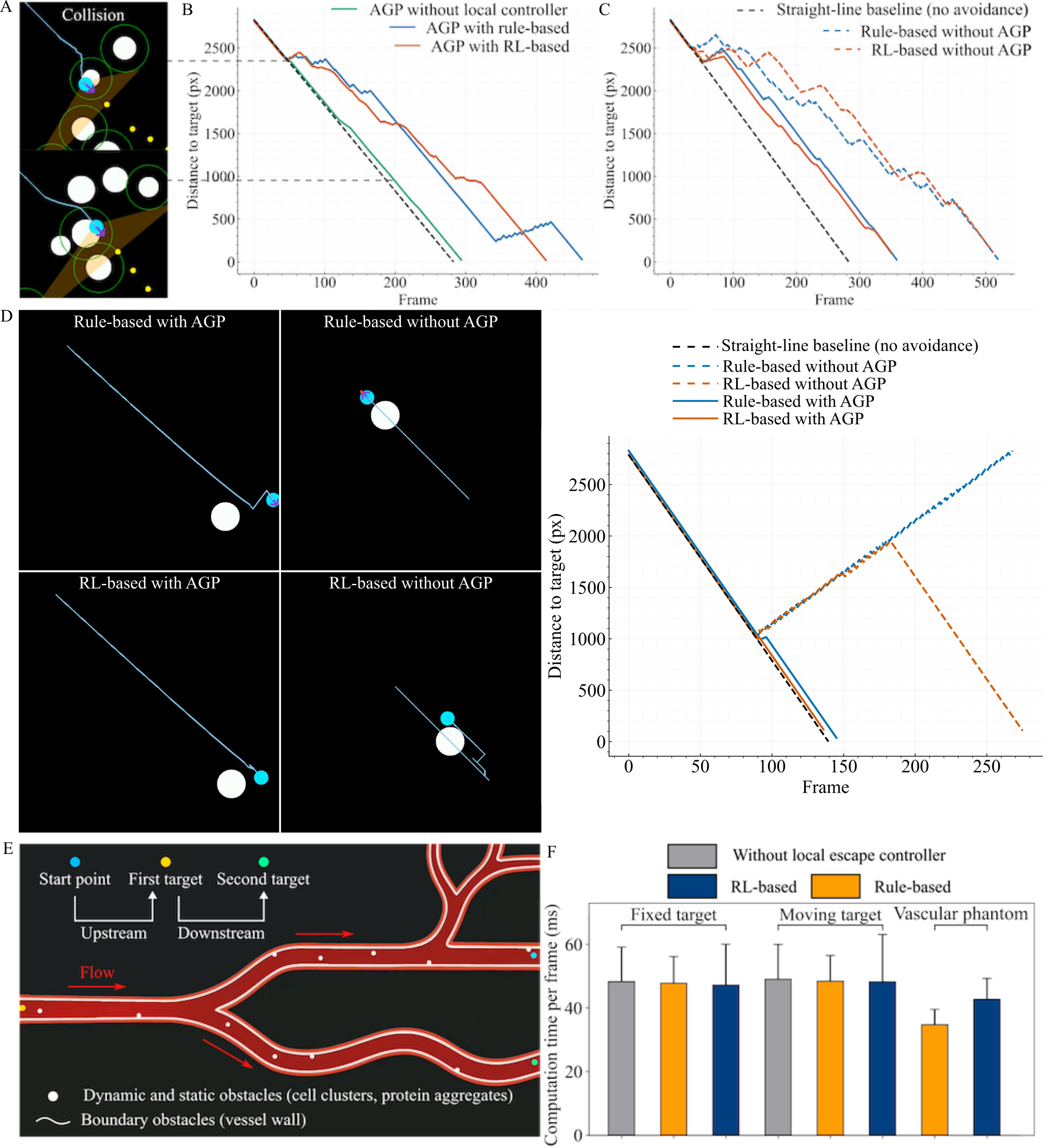}
    \caption{\textbf{Simulation evaluation of global--local dynamic navigation. See Supplementary Videos~5 and 6.} (\textbf{A}) Representative collision produced by AGP alone when a moving obstacle changes the locally safe geometry faster than global replanning can resolve the encounter. (\textbf{B}) Distance to a fixed target for AGP alone and AGP combined with the rule-based or RL-based local controller. (\textbf{C}) Effect of global guidance: local controllers used alone resolve short-range conflicts but can lose progress, whereas coupling them with AGP restores convergence toward the target. (\textbf{D}) Representative local-controller-only and AGP-coupled trajectories for an isolated moving-obstacle interaction, together with the corresponding target-distance histories. (\textbf{E}) Vascular-like environment containing boundary, static, and moving obstacles, with sequential upstream and downstream targets. (\textbf{F}) Per-frame computation time across the evaluated dynamic scenarios and controller configurations.}
    \label{fig:local_simulation}
\end{figure}

However, geometric replanning from the current obstacle configuration does not explicitly account for short-horizon obstacle motion. AGP alone could therefore still collide with a fast obstacle even when the route was recomputed at every step (Figure~\ref{fig:local_simulation}(A)). This behavior exposes the distinction between snapshot-level geometric feasibility and short-horizon dynamic safety. When either local controller was added, the microrobot could resolve the immediate encounter and continue toward the fixed target. The target-distance histories in Figure~\ref{fig:local_simulation}(B) show that both global--local configurations reached the goal in the representative trial. The learned controller generally produced fewer prolonged plateaus, whereas the rule-based controller occasionally moved temporarily away from the goal because its reflected escape point is determined only by local geometry.

The global planner was also necessary for sustained progress. When the local controllers were run without AGP, they could escape individual conflicts but had no representation of the distant objective, leading to unnecessary detours, plateaus, or motion away from the target. Reintroducing AGP restored global guidance for both local policies (Figure~\ref{fig:local_simulation}(C,D)). The same architecture was evaluated in the vascular-like simulation in Figure~\ref{fig:local_simulation}(E), where vessel walls were represented as boundary obstacles and moving circles represented flowing cells or aggregates. The global planner guided the microrobot between upstream and downstream targets, while the local controller handled nearby moving obstacles without interrupting overall progress toward the targets.

The added reactivity did not remove the real-time computational advantage of the framework. Across the fixed-target, moving-target, and vascular-like scenarios, per-frame computation remained below $50\,\mathrm{ms}$ (Figure~\ref{fig:local_simulation}(F)), which is compatible with the image-update rates used in the benchtop experiments.

\subsection*{Experimental autonomous navigation}

\begin{figure}
    \centering
    \includegraphics[width=0.75\textwidth]{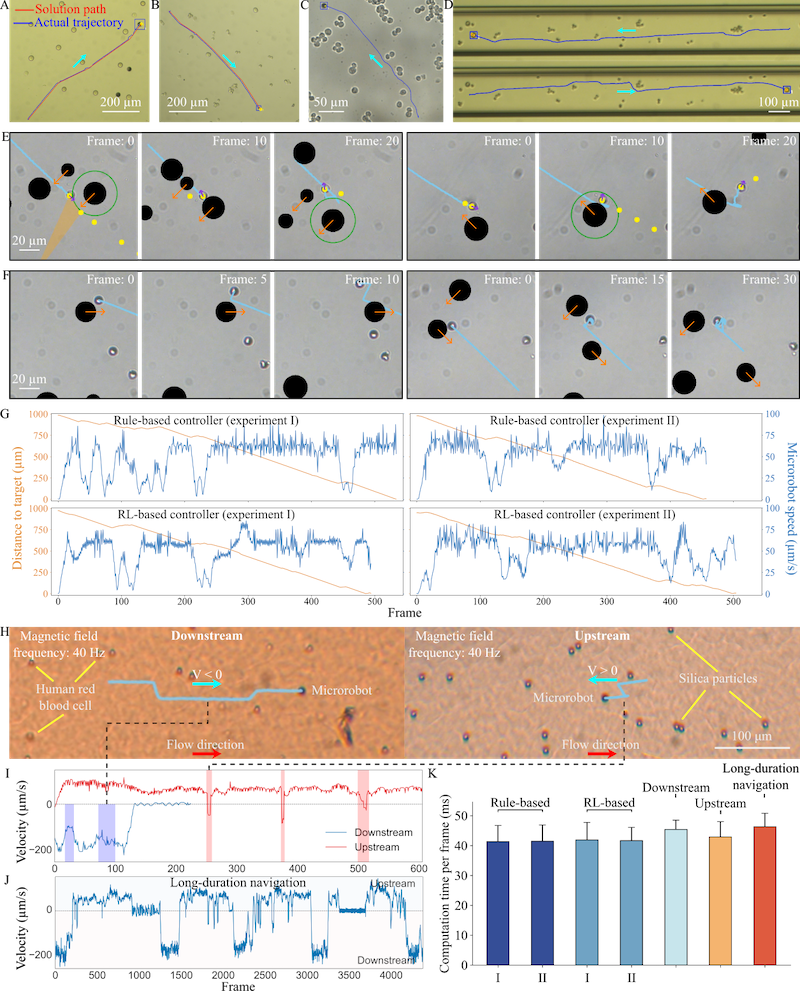}
    \caption{\textbf{Experimental closed-loop microrobot navigation.} (\textbf{A--D}) AGP navigation in static silica-particle and CHO-cell environments and inside a microfluidic channel, showing planned paths and measured microrobot trajectories. (\textbf{E}) Representative dynamic-avoidance sequences using the rule-based local controller. (\textbf{F}) Representative sequences using the RL-based local controller. (\textbf{G}) Distance to the target and estimated microrobot speed during dynamic avoidance. (\textbf{H}) Downstream navigation among diluted human red blood cells and upstream navigation among moving silica particles under controlled flow. (\textbf{I}) Corresponding velocity histories and local avoidance intervals. (\textbf{J}) Long-duration experiment with repeated upstream and downstream target transitions. (\textbf{K}) Per-frame computation time across the experimental conditions.}
    \label{fig:experiment}
\end{figure}

We next integrated the planner and local controllers with the magnetic actuation and microscopy platform. Static experiments first tested whether image-derived AGP routes could be executed by the physical microrobot. In silica-particle and CHO-cell environments, the measured trajectories closely followed the planned routes while maintaining separation from the detected obstacles (Figure~\ref{fig:experiment}(A,B); Supplementary Video~1). AGP was further validated across increasing obstacle densities, experimental silica-particle and CHO-cell fields, vascular geometries, and a maze-like environment (Supplementary Fig.~\ref{fig:AGP_static_additional}).

AGP was recomputed from the live image stream, so the path evolved with the measured microrobot state rather than being executed as a fixed open-loop trajectory (Figure~\ref{fig:experiment}(C)). Additional single-target and sequential multi-target experiments in a CHO-cell environment further demonstrated continuous online AGP replanning during closed-loop navigation (Supplementary Fig.~\ref{fig:AGP_online_experiment}; Supplementary Video~3). Navigation was also demonstrated inside a $200\,\mu\mathrm{m}$-wide microfluidic channel without imposed flow (Figure~\ref{fig:experiment}(D)).

Dynamic experiments then activated the local escape module. Representative rule-based and RL-based avoidance events are shown in Figure~\ref{fig:experiment}(E,F) and Supplementary Video~7. In both cases the microrobot was redirected around moving obstacles and subsequently resumed navigation toward the target. The corresponding target-distance and estimated-speed histories (Figure~\ref{fig:experiment}(G)) show repeated short local perturbations superimposed on overall progress toward the target. These experiments verify that the global--local switch can operate with real imaging, actuation latency, tracking noise, and nonideal microrobot motion rather than only in simulation.

To test the framework under a sustained environmental disturbance, we introduced controlled flow and suspended biological or particle obstacles. Under downstream conditions, a $10\,\mu\mathrm{m}$ microrobot navigated through diluted human red blood cells; under upstream conditions, silica particles were used as moving obstacles while the microrobot rolled against the imposed flow (Figure~\ref{fig:experiment}(H) and Supplementary Video~9). The RL-based local controller was used for these demonstrations without retraining specifically for flow. The mean imposed flow speed was approximately $400\,\mu\mathrm{m}\,\mathrm{s}^{-1}$ and the rotating magnetic field was operated at $40\,\mathrm{Hz}$. The measured mean microrobot speeds were approximately $164\,\mu\mathrm{m}\,\mathrm{s}^{-1}$ downstream and $62\,\mu\mathrm{m}\,\mathrm{s}^{-1}$ upstream. The velocity histories in Figure~\ref{fig:experiment}(I) show the local avoidance intervals superimposed on the flow-biased motion.

Finally, a long-duration experiment repeatedly alternated upstream and downstream traversal through a flowing red-blood-cell environment (Supplementary Video~10). The microrobot reached multiple sequential targets in both directions while preserving stable closed-loop operation over the full run (Figure~\ref{fig:experiment}(J)). Across the rule-based, RL-based, downstream, upstream, and long-duration experiments, the measured processing and planning time remained within the video-rate operating budget (Figure~\ref{fig:experiment}(K)). Together, these results demonstrate a progression from static path execution to dynamic avoidance and finally to sustained autonomous navigation under flow.

\subsection*{Navigation in biologically relevant environments}

\begin{figure}
    \centering
    \includegraphics[width=0.75\linewidth]{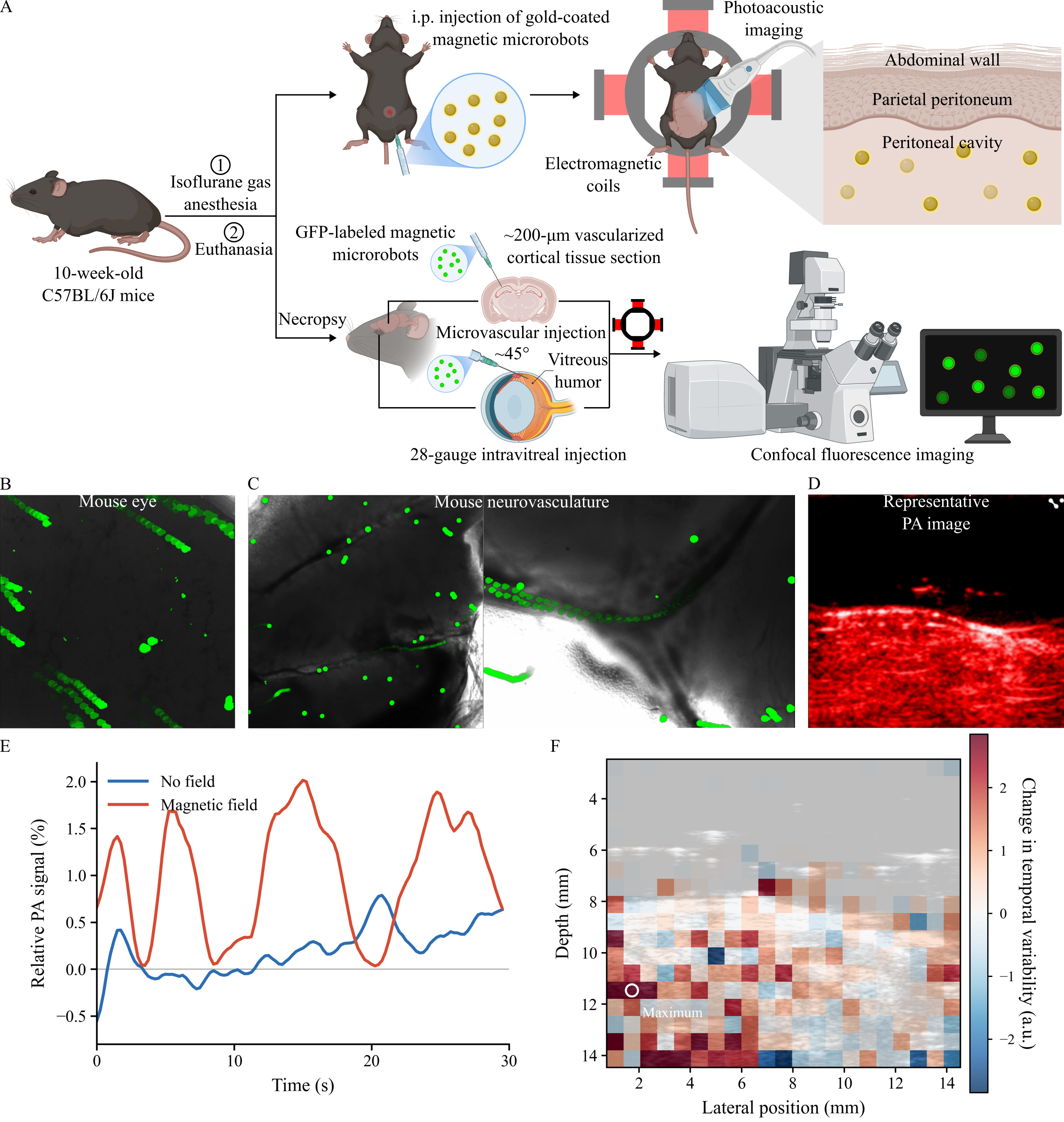}
    \caption{\textbf{Magnetic microrobot actuation and imaging in biologically relevant mouse tissues. See Supplementary Video~11.} \textbf{(A)} Workflow for post-mortem PAI and ex vivo eye and cortical-tissue experiments under rotating magnetic-field actuation. \textbf{(B,C)} Time-lapse fluorescence overlays showing GFP-labeled magnetic microrobot motion in the mouse eye and vascularized cortical tissue, respectively. \textbf{(D)} Representative intraperitoneal PA image of gold-coated magnetic microrobots. \textbf{(E)} Motion-corrected relative PA signal over matched 29.5-s no-field and magnetic-field intervals. \textbf{(F)} Field-associated change in temporal PA variability, $\Delta SD_{\mathrm{PA}}=SD_{\mathrm{field}}-SD_{\mathrm{no\,field}}$, overlaid on B-mode anatomy; the white circle marks the maximum increase.}
    \label{fig:biological}
\end{figure}

To evaluate magnetic microrobot actuation beyond synthetic microfluidic environments, we performed proof-of-feasibility experiments in post-mortem and ex vivo mouse tissues (Figure~\ref{fig:biological}). These experiments assessed whether magnetic microrobots could remain observable and magnetically actuated in biologically relevant environments. The experimental workflow is summarized in Figure~\ref{fig:biological}(A).

We first examined magnetic microrobot motion in ex vivo ocular and cerebral tissues. GFP-labeled magnetic microrobots were introduced into the vitreous humor of isolated mouse eyes and into the microvasculature of thin vascularized cortical tissue sections, respectively. Under rotating magnetic-field actuation, time-lapse fluorescence imaging revealed displacement of the microrobots within both tissue environments (Figure~\ref{fig:biological}(B,C)). These experiments demonstrate that magnetic actuation remains feasible in heterogeneous biological environments substantially more complex than the transparent microfluidic systems.

We next investigated photoacoustic imaging (PAI) as a sensing modality for magnetic microrobots when direct optical observation is limited. Gold-coated $20\,\mu\mathrm{m}$ magnetic microrobots were introduced into the intraperitoneal cavity of post-mortem mice and imaged at 750~nm while being actuated by a rotating magnetic field (Figure~\ref{fig:biological}(D)). Magnetic actuation produced a pronounced oscillatory PA response (Figure~\ref{fig:biological}(E)). Over matched 29.5-s intervals, the mean relative PA signal increased from 0.20\% without magnetic actuation to 1.00\% during magnetic actuation, while its temporal standard deviation increased from 0.28\% to 0.65\%, corresponding to a 2.33-fold increase in temporal variability. Spatial analysis further localized the field-associated increase in PA dynamics within the abdominal imaging region (Figure~\ref{fig:biological}(F)), with the maximum response observed at a lateral position of 1.72~mm and a depth of 11.47~mm.

Together, these experiments extend the platform beyond optically accessible microfluidic environments by demonstrating magnetic microrobot actuation in ocular and cerebral tissues and photoacoustic detection of magnetically induced dynamics in the intraperitoneal cavity. Although closed-loop AGP navigation was not performed in these biological settings, the results establish complementary actuation and sensing capabilities that could support future autonomous navigation in more realistic tissue environments.

\section*{DISCUSSION}

This work establishes a real-time autonomous navigation framework in which deterministic geometric planning and reactive local control are coupled to microscopy and magnetic actuation. The central design choice is the separation of roles across spatial and temporal scales. AGP continuously supplies a global route using the latest measured geometry, while local controllers intervene only when nearby obstacle motion makes the snapshot-level route temporarily unsafe. This modularity avoids requiring a learned policy to solve the complete navigation problem and preserves an interpretable global plan throughout execution.

The static benchmark clarifies the role of AGP within this architecture. Across the 50 paired environments, all four planners succeeded; AGP produced path lengths comparable to RRT* and COP while requiring substantially less planning time than the alternative implementations. The advantage arises from analytic obstacle pruning and local geometric construction rather than exhaustive raster search, stochastic tree expansion, or full geometry-graph construction.

Dynamic environments introduce a second limitation of purely geometric replanning. Even a collision-free route computed from the current image can become unsafe before the next control update when obstacles move rapidly. The local escape layer addresses this short-horizon mismatch. The rule-based controller provides a transparent and immediately deployable response, whereas the learned controller uses local position and velocity information to select context-dependent actions, including lateral, reverse, or stationary responses. The simulations and experiments show that local control alone is insufficient for reliable long-range navigation, but coupling it with AGP provides both reactivity and progress to the target.

The physical experiments extend this conclusion beyond idealized simulation. The framework operated with image-segmentation error, tracking noise, magnetic-actuation variability, finite control latency, channel boundaries, and imposed flow. The transition from static silica and CHO-cell scenes to dynamic obstacles, flowing red blood cells, and repeated upstream/downstream navigation is particularly important because it tests the closed-loop architecture under progressively less stationary conditions. The measured computation times remained compatible with video-rate operation, indicating that the computational structure of the framework is suitable for online microscopy-guided control rather than offline path generation.

The post-mortem and ex vivo experiments further probe whether the actuation and sensing components can be extended beyond optically simple microfluidic environments. GFP-labeled magnetic microrobots remained observable and magnetically mobile in the vitreous humor and vascularized cortical tissue, while gold-coated microrobots in the intraperitoneal cavity produced stronger and more variable PA signals during magnetic actuation than under the no-field condition. The spatial PA analysis localized the largest field-associated increase in temporal variability within the abdominal imaging region. Although closed-loop AGP navigation was not performed in tissue, these experiments demonstrate that magnetic actuation can be retained in heterogeneous biological environments and that PAI can report magnetically induced microrobot dynamics when direct optical observation is limited.

A significant gap nevertheless remains before in vivo autonomy can be achieved. The present planner depends on sufficiently reliable microrobot and obstacle localization; missed detections, occlusion, or segmentation latency can reduce the effective safety margin. Optical microscopy further constrains penetration depth and field of view, whereas the present PAI measurements provide sensing rather than full real-time state estimation for closed-loop planning. The biological experiments were also performed post-mortem or ex vivo and therefore do not reproduce physiological motion, perfusion, immune interactions, or other in vivo disturbances.

These limitations define the next steps toward autonomous microrobot operation in tissue. Deep-tissue sensing, including photoacoustic localization, must be integrated directly with the perception--planning--control loop; dynamic safety should explicitly account for obstacle motion and uncertainty; and the planning problem should ultimately extend to fully three-dimensional anatomy. Within those constraints, the present results show that fast analytic planning, local learned or rule-based reactivity, real-time magnetic control, and complementary biological sensing modalities can be combined as building blocks toward autonomous microrobot navigation in increasingly realistic environments.

\section*{MATERIALS AND METHODS}

\subsection*{Analytic geometry-based planner}

For each image frame, thresholding and contour extraction provide the position and geometry of the microrobot and surrounding obstacles. Let the center of obstacle $i$ be $\mathbf{c}_i=(x_{c,i},y_{c,i})^\top$, and let $w_i$ and $h_i$ denote the width and height of its axis-aligned bounding box. To account for the finite microrobot radius $R$, each obstacle is inflated into a circular safety region with
\begin{align}
    d_i &= \sqrt{w_i^2+h_i^2},\\
    r_i &= \frac{1}{2}d_i+R.
\end{align}
The microrobot can therefore be treated as a point moving among inflated obstacles.

AGP constructs a collision-free route from the start point $\mathcal{S}=(x_s,y_s)$ to the end point $\mathcal{E}=(x_e,y_e)$ using a sparse sequence of primary geometric nodes, followed by a denser waypoint sequence for closed-loop tracking. The planner is initialized from the straight-line path $\mathcal{S}\rightarrow\mathcal{E}$.

\subsubsection*{Obstacle pruning}

Define
\[
\mathbf{s}=(x_s,y_s)^\top,\qquad
\mathbf{e}=(x_e,y_e)^\top,\qquad
\mathbf{d}=\mathbf{e}-\mathbf{s}.
\]
The ideal path is the parametric line $\mathbf{s}+\tau\mathbf{d}$, $\tau\in\mathbb{R}$. For each obstacle center $\mathbf{c}_i$, its projection onto the ideal path is
\begin{equation}
    t_i=
    \frac{(\mathbf{c}_i-\mathbf{s})\cdot\mathbf{d}}
    {\|\mathbf{d}\|^2}.
\end{equation}
Only obstacles satisfying $0\leq t_i\leq1$ are retained at this stage, such that the orthogonal foot
\begin{equation}
    \mathbf{g}_i=\mathbf{s}+t_i\mathbf{d}
\end{equation}
lies between $\mathcal{S}$ and $\mathcal{E}$. Among these obstacles, AGP discards those whose center-to-foot distance exceeds a scaled safety region,
\begin{equation}
    \|\mathbf{c}_i-\mathbf{g}_i\|>\alpha r_i
    \quad\Rightarrow\quad
    \text{discard},
    \qquad \alpha>0.
\end{equation}
The remaining obstacles are used for subsequent geometric path construction.

\subsubsection*{Primary-node construction}

The retained obstacle centers, projected foot points, and safety regions are ordered by increasing $x$-coordinate of the projected foot points, $x_{g,i}$. For a nonvertical ideal path, its slope is
\begin{equation}
    k_o=\frac{y_e-y_s}{x_e-x_s}.
\end{equation}
The family of lines passing through each obstacle center and perpendicular to the ideal path has slope
\begin{equation}
    k_p=-\frac{1}{k_o}
\end{equation}
and intercept
\begin{equation}
    e_{p,i}=y_{c,i}-k_p x_{c,i}.
\end{equation}
Each primary node $v_i=(x_{v,i},y_{v,i})^\top$ is constrained to the corresponding perpendicular line,
\begin{equation}
    y_{v,i}=k_p x_{v,i}+e_{p,i}.
\end{equation}
For a vertical ideal path ($x_e=x_s$), the equivalent vector constraint
\begin{equation}
    (v_i-\mathbf{c}_i)\cdot\mathbf{d}=0
\end{equation}
is used.

Each candidate node is required to remain outside the relevant safety regions,
\begin{equation}
    r_i^2\leq\|v_j-\mathbf{c}_i\|^2.
\end{equation}

For each current node $v_i$, AGP constructs tangents to the two nearest active safety regions. Define
\begin{equation}
    \theta_i=
    \operatorname{atan2}
    (y_{c,i}-y_{v,i},\,x_{c,i}-x_{v,i}),
    \qquad
    \rho_i=\|\mathbf{c}_i-v_i\|,
\end{equation}
with $\rho_i>r_i$. The two tangent slopes are
\begin{equation}
    m_{1,2}
    =
    \tan\!\left(
    \theta_i
    \pm
    \arcsin\!\frac{r_i}{\rho_i}
    \right).
\end{equation}

Together with the line through the next projected foot point $\mathbf{g}_{i+1}$ that is parallel to the perpendicular family, these tangents define the local geometric construction used to determine the next primary node. AGP first projects the current node $v_i$ perpendicularly onto the corresponding side of this construction. If the resulting point lies outside all active safety regions, it is accepted as $v_{i+1}$. Otherwise, $v_{i+1}$ is selected as the intersection of the shortest valid tangent with that side.

After each node update, the obstacle closest to the current node is removed from the active set, and the construction is repeated using the two obstacles nearest the new node. If only one obstacle remains, the shorter valid tangent is used. If the straight segment from the current node to $\mathcal{E}$ has at most one intersection with the remaining safety region, the end point is assigned directly as the next node.

\subsubsection*{Noncircular obstacle representation}

Elongated structures and vessel boundaries are represented by discretizing the detected bounding geometry into square cells of side length $g$. Cells containing contour pixels are retained and grouped into local sub-contours. Each retained cell is assigned a circular safety region with
\begin{equation}
    d_{\mathrm{cell}}=\sqrt{2}\,g,
    \qquad
    r_{\mathrm{cell}}=
    \frac{1}{2}d_{\mathrm{cell}}+R.
\end{equation}
Representing an extended structure by multiple local safety regions, rather than a single enclosing region, preserves additional free space near curved boundaries while allowing the same AGP construction to be applied.

\subsubsection*{Waypoint generation}

After all primary nodes have been determined, additional waypoints are inserted along each segment in proportion to its length. The resulting denser path is used as the reference trajectory for closed-loop microrobot tracking.

\subsection*{Planner benchmarking}

AGP was benchmarked against weighted A*, RRT*, and circular-obstacle pathfinding (COP). Fifty independently seeded environments were generated, each containing 20 circular obstacles. For each seed, all planners received the same obstacle realization and fixed start--goal pair.

Performance was quantified using geometric path length, cumulative absolute turning angle, and wall-clock planning time. For a polyline
$P=\{\mathbf{p}_0,\ldots,\mathbf{p}_M\}$,
the path length was defined as
\begin{equation}
    L(P)=\sum_{k=1}^{M}\|\mathbf{p}_k-\mathbf{p}_{k-1}\|_2,
\end{equation}
and path complexity was quantified by the accumulated absolute heading change,
\begin{equation}
    \Theta(P)=
    \sum_{k=1}^{M-1}
    \left|
    \operatorname{wrap}_{[-\pi,\pi)}
    (\psi_k-\psi_{k-1})
    \right|,
\end{equation}
where $\psi_k$ denotes the heading of segment $k$. Planner-specific implementation details and hyperparameters are provided in the Supplementary Information.

\subsection*{Local obstacle-avoidance controllers}

\subsubsection*{Rule-based local escape controller}

Let $\mathbf{m}$ denote the microrobot center, $\mathbf{a}$ the tip of a one-step look-ahead vector, and $\mathbf{c}_i$ the center of obstacle $i$. The look-ahead length is
\begin{equation}
    L_a=v_m+v_{c,\max},
    \qquad
    v_{c,\max}=\max_i v_{c,i},
\end{equation}
where $v_m$ and $v_{c,i}$ denote the microrobot and obstacle displacements over one control step, respectively.

Two geometric cases define the temporary escape target. In Case~I, the look-ahead point $\mathbf{a}$ enters an obstacle safety region. The point is reflected across the microrobot center,
\begin{equation}
    \mathbf{a}'=2\mathbf{m}-\mathbf{a},
\end{equation}
and $\mathbf{a}'$ is used as the temporary escape target. In Case~II, the microrobot center itself enters the safety region. The obstacle center is reflected across $\mathbf{m}$,
\begin{equation}
    \mathbf{c}'_i=2\mathbf{m}-\mathbf{c}_i.
\end{equation}
The local geometry is re-evaluated at every control step so that the temporary target updates with obstacle motion. Once the conflict is cleared, the temporary command is released and control returns to the current AGP route.

\subsubsection*{RL-based local escape controller}

The RL-based controller was trained using deep Q-learning to resolve short-range obstacle interactions independently of the global target or AGP path \cite{mnih2015human}. For obstacle $i$, the local safety-region radius was defined as
\begin{equation}
    r_i^{\mathrm{safe}}=r_i+R+\delta,
\end{equation}
where $r_i$ and $R$ are the physical radii of the obstacle and microrobot, respectively. The one-step relative-motion buffer was
\begin{equation}
    \delta=v_m+v_{c,\max},
    \qquad
    v_{c,\max}=\max_i v_i,
\end{equation}
where $v_m$ and $v_i$ denote microrobot and obstacle displacements over one simulation step. The policy was queried only when the microrobot entered at least one local safety region.

At each local control step, the controller selected from nine discrete actions: one stationary action and eight uniformly spaced planar motion directions. The observation consisted of the normalized relative positions and velocities of the four nearest obstacles,
\begin{equation}
    \mathbf{o}=
    \left[
    \mathbf{o}_1^\top,
    \mathbf{o}_2^\top,
    \mathbf{o}_3^\top,
    \mathbf{o}_4^\top
    \right]^\top
    \in\mathbb{R}^{16},
\end{equation}
with
\begin{equation}
    \mathbf{o}_i=
    \left[
    \frac{\Delta x_i}{D_{\max}},
    \frac{\Delta y_i}{D_{\max}},
    \frac{v_{x,i}}{v_{c,\max}},
    \frac{v_{y,i}}{v_{c,\max}}
    \right]^\top,
    \qquad
    D_{\max}=\sqrt{W^2+H^2}.
\end{equation}
If fewer than four obstacles were available, the unused observation entries were zero padded.

The local safety state was described by the minimum signed clearance
\begin{equation}
    \phi_t=
    \min_i
    \left(
    \|\mathbf{m}_t-\mathbf{c}_{i,t}\|_2-r_i^{\mathrm{safe}}
    \right),
\end{equation}
such that $\phi_t<0$ indicates that the microrobot lies inside at least one local safety region. The reward was defined as
\begin{equation}
    r_t=
    \begin{cases}
        R_{\mathrm{collision}},
        & \exists i:\ \|\mathbf{m}_t-\mathbf{c}_{i,t}\|_2<r_i+R,\\
        R_{\mathrm{success}},
        & \phi_{t-1}<0\ \land\ \phi_t\geq0,\\
        k_{\mathrm{shape}}(\phi_t-\phi_{t-1})-\lambda_{\mathrm{step}},
        & \phi_t<0,\\
        0,
        & \phi_t\geq0.
    \end{cases}
\end{equation}
The complete reward parameters are listed in Supplementary Table~S2.

\subsubsection*{DQN training and implementation}

The DQN was trained in a $2000\times2000$-pixel workspace containing 50 dynamic and five static circular obstacles. Each episode began with the microrobot near the center of the workspace and an initial travel direction sampled uniformly from eight planar directions, emulating a local segment of motion before a safety-region encounter. Dynamic obstacles moved along one of eight discretized directions and reflected from workspace boundaries and other obstacles. The complete environment and motion parameters are summarized in Supplementary Table~S1.

The Q-network contained two fully connected hidden layers with 512 and 256 units and rectified linear unit activations, followed by nine output values corresponding to the discrete action set. Training used $\varepsilon$-greedy exploration, experience replay, and periodic target-network synchronization. The complete DQN training configuration is provided in Supplementary Table~S3.

Policy performance was evaluated periodically in an independent validation environment. A checkpoint was retained whenever validation success exceeded the previous best value, with mean return used to break ties. The selected policy was subsequently used in the dynamic-navigation simulations and benchtop experiments without retraining.

\subsection*{Experimental platform and navigation protocols}

\begin{figure}
    \centering
    \includegraphics[width=0.75\textwidth]{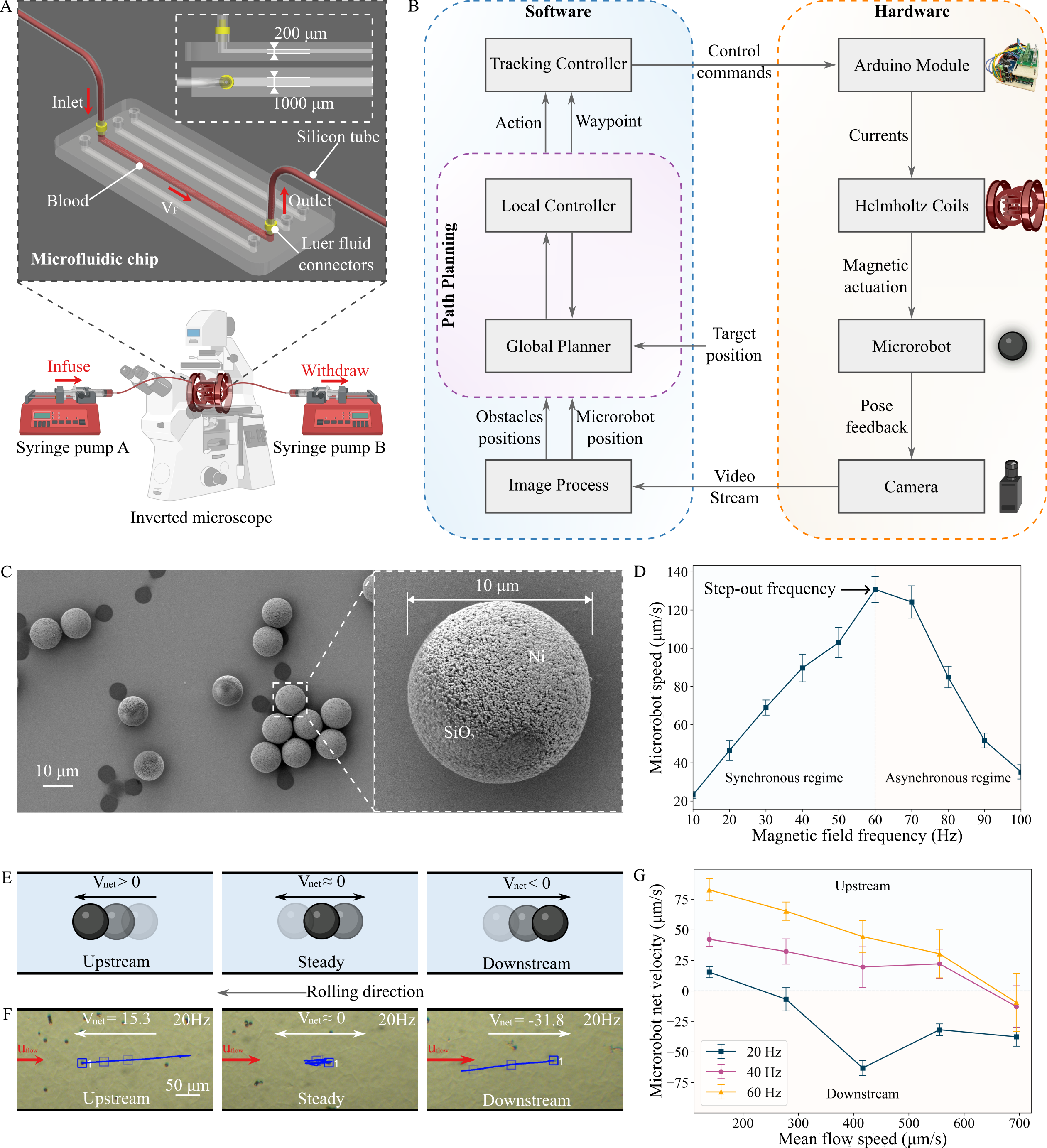}
    \caption{\textbf{Experimental platform and actuation characterization.} (\textbf{A}) Controlled-flow microfluidic platform integrated with an inverted microscope and three-axis Helmholtz coil system. (\textbf{B}) Closed-loop software--hardware architecture linking camera feedback, image processing, global planning, local control, tracking control, coil actuation, and the microrobot. (\textbf{C}) Scanning-electron-microscopy image of nickel-coated silica microrobots. (\textbf{D}) Microrobot rolling speed versus rotating-field frequency, showing synchronous operation and step-out. (\textbf{E}) Schematic definition of upstream, steady, and downstream motion. (\textbf{F}) Representative bright-field trajectories under the three flow regimes. (\textbf{G}) Mean microrobot net velocity as a function of mean flow speed at different actuation frequencies.}
    \label{fig:setup}
\end{figure}

A straight-channel microfluidic device was connected to two programmable syringe pumps through silicone tubing and Luer fittings. One pump infused the working fluid while the second withdrew at the same nominal rate to establish controlled flow through the observation region. The chip was positioned within a three-axis Helmholtz coil set mounted on an inverted microscope for real-time optical observation (Figure~\ref{fig:setup}(A); Supplementary Table~S4). A photograph of the complete benchtop experimental platform, integrating microscopy, magnetic actuation, microfluidic flow control, and closed-loop navigation, is provided in Supplementary Fig.~\ref{fig:experimental_platform}.

Figure~\ref{fig:setup}(B) summarizes the closed-loop software--hardware architecture. Camera images were processed to estimate the microrobot and obstacle states. The global planner generated the current waypoint sequence, a local controller could temporarily modify the commanded motion during short-range conflicts, and a tracking controller converted the resulting command into magnetic actuation. An Arduino-based driver system supplied current to the orthogonal Helmholtz coils, and camera-based position feedback closed the control loop. The magnetic-field geometry and waypoint-switching rule are shown in Supplementary Fig.~\ref{fig:control}.

Magnetic microrobots were fabricated by coating commercial $10\,\mu\mathrm{m}$ silica beads with a nickel half shell by electron-beam physical vapor deposition. A representative scanning-electron-microscopy image is shown in Figure~\ref{fig:setup}(C). Rolling performance was characterized by measuring microrobot speed as a function of rotating-field frequency (Figure~\ref{fig:setup}(D)). The interaction between magnetic rolling and background flow was evaluated in upstream, steady, and downstream regimes (Figure~\ref{fig:setup}(E--G) and Supplementary Video~8). The net axial velocity was described by
\begin{equation}
    V_{\mathrm{net}}=v_{\mathrm{roll}}-u_{\mathrm{flow}},
    \qquad
    v_{\mathrm{roll}}=\alpha_{\mathrm{roll}}(2\pi R f),
\end{equation}
where $u_{\mathrm{flow}}$ is the mean flow speed, $f$ is the rotating-field frequency, $R$ is the microrobot radius, and $\alpha_{\mathrm{roll}}<1$ accounts for wall coupling, hydrodynamic slip, and imperfect rolling.

For the dynamic-obstacle benchtop experiments, a $10\,\mu\mathrm{m}$ microrobot was observed under a $10\times$ objective and driven with a $35\,\mathrm{Hz}$ rotating magnetic field while black circular objects served as moving obstacles. For the flow experiments, the mean imposed flow speed was approximately $400\,\mu\mathrm{m}\,\mathrm{s}^{-1}$ and the rotating magnetic field was $40\,\mathrm{Hz}$. Upstream trials used $10\,\mu\mathrm{m}$ silica particles as moving obstacles, whereas downstream trials used diluted human red blood cells. The RL-based local controller was transferred to the flow experiments without additional training.

\subsection*{Image processing}

Each incoming video frame was converted to grayscale for contour extraction; HSV representations were additionally used when color-based discrimination was required. Thresholded masks were used to identify the microrobot, obstacles, and relevant workspace boundaries, followed by contour extraction to obtain their geometric representations. For each detected obstacle, an axis-aligned bounding box provided its center and dimensions. These image-derived geometries were then used to construct the safety regions defined in the Results. Supplementary Fig.~\ref{fig:safety} shows the complete image-processing workflow from raw fluorescence or bright-field images to binary masks, contour extraction, and geometric state estimation.

\subsection*{Blood preparation and cell culture}

Whole blood was obtained from \textsc{STEMCELL} Technologies (catalog 70508.2; donor RB1910; acquired 20 March 2025) and diluted 1:40 in Dulbecco's phosphate-buffered saline (DPBS; Gibco, 14040133) for the reported red-blood-cell experiments.

Chinese hamster ovary (CHO) cells were provided by collaborators and maintained in DMEM/F-12 supplemented with 10\% fetal bovine serum and 1\% penicillin--streptomycin under standard culture conditions. At approximately 85\% confluence, cells were washed with DPBS and detached using TrypLE Express. Cells between passages 3 and 8 were used.

\subsection*{Mouse and ex vivo tissue preparation}

All animal procedures were performed in accordance with an approved Institutional Animal Care and Use Committee (IACUC) protocol 1443-2025-0. Ten-week-old C57BL/6J mice were used for the post-mortem and ex vivo experiments. Animals were anesthetized with isoflurane and euthanized according to the approved protocol before tissue harvesting or post-mortem imaging. The overall experimental workflow is illustrated in Figure~\ref{fig:biological}(A).

For the ex vivo experiments, eyes and brains were harvested post-mortem and fixed in 4\% paraformaldehyde (PFA) in phosphate-buffered saline (PBS) at $4\,^{\circ}\mathrm{C}$ for 24 h. For the cerebral-tissue experiments, approximately $200\,\mu\mathrm{m}$-thick sections of vascularized cortical tissue were manually isolated from the brain surface, and GFP-labeled magnetic microrobots were introduced into the microvasculature using a 28-gauge needle. For the ocular experiments, GFP-labeled magnetic microrobots were injected into the vitreous humor using a 28-gauge needle inserted at an approximately $45^{\circ}$ angle. The prepared samples were positioned on a fluorescence microscope equipped with a magnetic actuation system. A rotating magnetic field was applied to actuate the microrobots, and their motion within the tissue samples was recorded by fluorescence microscopy.

For the post-mortem photoacoustic imaging experiments, abdominal hair was removed using a razor followed by depilatory cream to improve optical and acoustic access. Gold-coated $20\,\mu\mathrm{m}$ magnetic microrobots were injected into the intraperitoneal cavity, and acoustic coupling was established by applying ultrasound gel to the abdominal surface. A magnetic coil system integrated with the photoacoustic imaging setup generated a rotating magnetic field during image acquisition. Before the animal experiments, the wavelength-dependent PA response of the gold-coated microrobots was characterized using separate tubes containing deionized water and a suspension of the microrobots in deionized water. Based on this characterization, 750~nm was selected as the imaging wavelength to provide strong microrobot contrast relative to the water background.

The recorded PA sequences were analyzed at an effective cine rate of 4~Hz. Translational tissue motion was estimated from the corresponding B-mode images using phase-correlation registration and corrected before PA quantification. The PA signal was extracted from the red image channel as
\begin{equation}
    I_{\mathrm{PA}}=\max\left[R-\frac{G+B}{2},0\right],
\end{equation}
and spatially averaged within a fixed imaging region. The complete 29.5-s no-field recording was compared with an equal-duration magnetic-field interval. Relative PA signal changes were calculated as
\begin{equation}
    \Delta I_{\mathrm{PA}}(t)=100\left[\frac{I(t)}{I_0}-1\right],
\end{equation}
where $I_0$ was the median signal during the initial 5~s of each acquisition. A 1-s centered moving average was applied for visualization.

For spatial analysis, the imaging region was divided into approximately $0.69\times0.67$~mm bins. Magnetic-field-associated PA dynamics were quantified as
\begin{equation}
    \Delta SD_{\mathrm{PA}}=SD_{\mathrm{field}}-SD_{\mathrm{no\,field}},
\end{equation}
where $SD_{\mathrm{field}}$ and $SD_{\mathrm{no\,field}}$ denote the temporal standard deviations of the PA signal within each spatial bin under the magnetic-field and no-field conditions, respectively.


\clearpage
\bibliography{reference}

@article{jiang2022control,
	author = {Jiang, Jialin and Yang, Zhengxin and Ferreira, Antoine and Zhang, Li},
	journal = {Advanced Intelligent Systems},
	number = {5},
	pages = {2100279},
	publisher = {Wiley Online Library},
	title = {Control and autonomy of microrobots: Recent progress and perspective},
	volume = {4},
	year = {2022}}

@article{yang2019automated,
	author = {Yang, Lidong and Zhang, Yabin and Wang, Qianqian and Chan, Kai-Fung and Zhang, Li},
	journal = {IEEE Transactions on Automation Science and Engineering},
	number = {1},
	pages = {490--501},
	publisher = {IEEE},
	title = {Automated control of magnetic spore-based microrobot using fluorescence imaging for targeted delivery with cellular resolution},
	volume = {17},
	year = {2019}}

@article{ceylan2019translational,
	author = {Ceylan, Hakan and Yasa, Immihan C and Kilic, Ugur and Hu, Wenqi and Sitti, Metin},
	journal = {Progress in Biomedical Engineering},
	number = {1},
	pages = {012002},
	publisher = {IOP Publishing},
	title = {Translational prospects of untethered medical microrobots},
	volume = {1},
	year = {2019}}

@article{aziz2020medical,
	author = {Aziz, Azaam and Pane, Stefano and Iacovacci, Veronica and Koukourakis, Nektarios and Czarske, J{\"u}rgen and Menciassi, Arianna and Medina-S{\'a}nchez, Mariana and Schmidt, Oliver G},
	journal = {ACS nano},
	number = {9},
	pages = {10865--10893},
	publisher = {ACS Publications},
	title = {Medical imaging of microrobots: Toward in vivo applications},
	volume = {14},
	year = {2020}}

@article{nelson2010microrobots,
	author = {Nelson, Bradley J and Kaliakatsos, Ioannis K and Abbott, Jake J},
	journal = {Annual review of biomedical engineering},
	pages = {55--85},
	publisher = {Annual Reviews},
	title = {Microrobots for minimally invasive medicine},
	volume = {12},
	year = {2010}}

@article{li2018development,
	author = {Li, Junyang and Li, Xiaojian and Luo, Tao and Wang, Ran and Liu, Chichi and Chen, Shuxun and Li, Dongfang and Yue, Jianbo and Cheng, Shuk-han and Sun, DJSR},
	journal = {Science robotics},
	number = {19},
	pages = {eaat8829},
	publisher = {American Association for the Advancement of Science},
	title = {Development of a magnetic microrobot for carrying and delivering targeted cells},
	volume = {3},
	year = {2018}}

@article{jeon2019magnetically,
	author = {Jeon, Sungwoong and Kim, Sangwon and Ha, Shinwon and Lee, Seungmin and Kim, Eunhee and Kim, So Yeun and Park, Sun Hwa and Jeon, Jung Ho and Kim, Sung Won and Moon, Cheil and others},
	journal = {Science Robotics},
	number = {30},
	pages = {eaav4317},
	publisher = {American Association for the Advancement of Science},
	title = {Magnetically actuated microrobots as a platform for stem cell transplantation},
	volume = {4},
	year = {2019}}

@article{fan2022obstacle,
	author = {Fan, Qigao and Cui, Guangming and Zhao, Zhengqing and Shen, Jun},
	journal = {IEEE Robotics and Automation Letters},
	number = {4},
	pages = {9794--9801},
	publisher = {IEEE},
	title = {Obstacle avoidance for microrobots in simulated vascular environment based on combined path planning},
	volume = {7},
	year = {2022}}

@article{salehizadeh2021path,
	author = {Salehizadeh, Mohammad and Diller, Eric D},
	journal = {IEEE robotics and automation letters},
	number = {2},
	pages = {2674--2681},
	publisher = {IEEE},
	title = {Path planning and tracking for an underactuated two-microrobot system},
	volume = {6},
	year = {2021}}

@article{zou2022dynamic,
	author = {Zou, Qian and Du, Xingzhou and Liu, Yuezhen and Chen, Hui and Wang, Yibin and Yu, Jiangfan},
	journal = {IEEE Transactions on Automation Science and Engineering},
	publisher = {IEEE},
	title = {Dynamic path planning and motion control of microrobotic swarms for mobile target tracking},
	year = {2022}}

@article{wang2021micromanipulation,
	author = {Wang, Qianqian and Yang, Lidong and Zhang, Li},
	journal = {IEEE Transactions on Automation Science and Engineering},
	number = {2},
	pages = {759--771},
	publisher = {IEEE},
	title = {Micromanipulation using reconfigurable self-assembled magnetic droplets with needle guidance},
	volume = {19},
	year = {2021}}

@inproceedings{yang2023rolling,
  title={Rolling Helical Microrobots for Cell Patterning},
  author={Yang, Yanda and Kirmizitas, Fatma Ceren and Sokolich, Max and Valencia, Alejandra and Rivas, David and Karakan, M {\c{C}}a{\u{g}}atay and White, Alice E and Malikopoulos, Andreas A and Das, Sambeeta},
  booktitle={2023 International Conference on Manipulation, Automation and Robotics at Small Scales (MARSS)},
  pages={1--6},
  year={2023},
  organization={IEEE}
}

@inproceedings{yang2023closed,
  title={Closed-Loop Control of Bubble-Propelled Microrobots},
  author={Yang, Yanda and Rivas, David and Sokolich, Max and Das, Sambeeta},
  booktitle={2023 International Conference on Manipulation, Automation and Robotics at Small Scales (MARSS)},
  pages={1--6},
  year={2023},
  organization={IEEE}
}

@article{yang2024quadrupole,
  title={Quadrupole Magnetic Tweezers for Precise Cell Transportation},
  author={Yang, Yanda and Sokolich, Max and Mallick, Sudipta and Das, Sambeeta},
  journal={IEEE Transactions on Biomedical Engineering},
  year={2024},
  publisher={IEEE}
}

@article{yang2021autonomous,
  title={Autonomous navigation of magnetic microrobots in a large workspace using mobile-coil system},
  author={Yang, Zhengxin and Yang, Lidong and Zhang, Li},
  journal={IEEE/ASME Transactions on Mechatronics},
  volume={26},
  number={6},
  pages={3163--3174},
  year={2021},
  publisher={IEEE}
}

@article{liu2019navigation,
  title={Navigation and visual feedback control for magnetically driven helical miniature swimmers},
  author={Liu, Jia and Xu, Tiantian and Yang, Simon X and Wu, Xinyu},
  journal={IEEE Transactions on Industrial Informatics},
  volume={16},
  number={1},
  pages={477--487},
  year={2019},
  publisher={IEEE}
}

@article{kim2017autonomous,
  title={Autonomous dynamic obstacle avoidance for bacteria-powered microrobots (BPMs) with modified vector field histogram},
  author={Kim, Hoyeon and Cheang, U Kei and Kim, Min Jun},
  journal={PloS one},
  volume={12},
  number={10},
  pages={e0185744},
  year={2017},
  publisher={Public Library of Science San Francisco, CA USA}
}

@article{li2017autonomous,
  title={Autonomous collision-free navigation of microvehicles in complex and dynamically changing environments},
  author={Li, Tianlong and Chang, Xiaocong and Wu, Zhiguang and Li, Jinxing and Shao, Guangbin and Deng, Xinghong and Qiu, Jianbin and Guo, Bin and Zhang, Guangyu and He, Qiang and others},
  journal={Acs Nano},
  volume={11},
  number={9},
  pages={9268--9275},
  year={2017},
  publisher={ACS Publications}
}

@article{yang2022hierarchical,
  title={Hierarchical planning with deep reinforcement learning for 3D navigation of microrobots in blood vessels},
  author={Yang, Yuguang and Bevan, Michael A and Li, Bo},
  journal={Advanced Intelligent Systems},
  volume={4},
  number={11},
  pages={2200168},
  year={2022},
  publisher={Wiley Online Library}
}

@inproceedings{jiang2023dqn,
  title={DQN-based on-line path planning method for automatic navigation of miniature robots},
  author={Jiang, Jialin and Yang, Lidong and Zhang, Li},
  booktitle={2023 IEEE International Conference on Robotics and Automation (ICRA)},
  pages={5407--5413},
  year={2023},
  organization={IEEE}
}

@article{yang2022autonomous,
  title={Autonomous environment-adaptive microrobot swarm navigation enabled by deep learning-based real-time distribution planning},
  author={Yang, Lidong and Jiang, Jialin and Gao, Xiaojie and Wang, Qinglong and Dou, Qi and Zhang, Li},
  journal={Nature Machine Intelligence},
  volume={4},
  number={5},
  pages={480--493},
  year={2022},
  publisher={Nature Publishing Group UK London}
}

@article{yang2026residual,
  title={Residual RL-MPC for Robust Microrobotic Cell Pushing Under Time-Varying Flow},
  author={Yang, Yanda and Das, Sambeeta},
  journal={arXiv preprint arXiv:2603.05448},
  year={2026}
}

@article{yang2026micropush,
  title={MicroPush: A Simulator and Benchmark for Contact-Rich Cell Pushing and Assembly with a Magnetic Rolling Microrobot},
  author={Yang, Yanda and Das, Sambeeta},
  journal={arXiv preprint arXiv:2602.23607},
  year={2026}
}

@article{chowdhury2015controlling,
  title={Controlling multiple microrobots: recent progress and future challenges},
  author={Chowdhury, Sagar and Jing, Wuming and Cappelleri, David J},
  journal={Journal of Micro-Bio Robotics},
  volume={10},
  number={1},
  pages={1--11},
  year={2015},
  publisher={Springer}
}

@article{yang2020magnetic,
  title={Magnetic actuation systems for miniature robots: A review},
  author={Yang, Zhengxin and Zhang, Li},
  journal={Advanced Intelligent Systems},
  volume={2},
  number={9},
  pages={2000082},
  year={2020},
  publisher={Wiley Online Library}
}

@article{du2023image,
  title={Image-integrated magnetic actuation systems for localization and remote actuation of medical miniature robots: A survey},
  author={Du, Xingzhou and Yu, Jiangfan},
  journal={IEEE Transactions on Robotics},
  volume={39},
  number={4},
  pages={2549--2568},
  year={2023},
  publisher={IEEE}
}

@article{hou2023review,
  title={A review on microrobots driven by optical and magnetic fields},
  author={Hou, Yaozhen and Wang, Huaping and Fu, Rongxin and Wang, Xian and Yu, Jiangfan and Zhang, Shuailong and Huang, Qiang and Sun, Yu and Fukuda, Toshio},
  journal={Lab on a Chip},
  volume={23},
  number={5},
  pages={848--868},
  year={2023},
  publisher={The Royal Society of Chemistry}
}

@article{landers2025clinically,
  title={Clinically ready magnetic microrobots for targeted therapies},
  author={Landers, Fabian C and Hertle, Lukas and Pustovalov, Vitaly and Sivakumaran, Derick and Oral, Cagatay M and Brinkmann, Oliver and Meiners, Kirstin and Theiler, Pascal and Gantenbein, Valentin and Veciana, Andrea and others},
  journal={Science},
  volume={390},
  number={6774},
  pages={710--715},
  year={2025},
  publisher={American Association for the Advancement of Science}
}

@article{jia2021robust,
  title={Robust navigation control of a microrobot with hysteresis compensation},
  author={Jia, Yuanjun and Zheng, Liushuai and Dong, Dingran and Wang, Yong and Sun, Dong},
  journal={IEEE Transactions on Automation Science and Engineering},
  volume={19},
  number={4},
  pages={3083--3092},
  year={2021},
  publisher={IEEE}
}

@article{wang2024deep,
  title={Deep reinforcement learning-based collision-free navigation for magnetic helical microrobots in dynamic environments},
  author={Wang, Huaping and Qiu, Yukang and Hou, Yaozhen and Shi, Qing and Huang, Hen-Wei and Huang, Qiang and Fukuda, Toshio},
  journal={IEEE Transactions on Automation Science and Engineering},
  volume={22},
  pages={7810--7820},
  year={2024},
  publisher={IEEE}
}

@article{zhong2026adaptive,
  title={Adaptive shared cascade navigation control of magnetic microrobots in unstructured dynamic environments},
  author={Zhong, Shihao and Hou, Yaozhen and Zheng, Zhiqiang and Huang, Hen-Wei and Shi, Qing and Huang, Qiang and Fukuda, Toshio and Wang, Huaping},
  journal={IEEE Transactions on Cybernetics},
  year={2026},
  publisher={IEEE}
}

@article{pohl1970heuristic,
  title={Heuristic search viewed as path finding in a graph},
  author={Pohl, Ira},
  journal={Artificial intelligence},
  volume={1},
  number={3-4},
  pages={193--204},
  year={1970},
  publisher={Elsevier}
}

@article{karaman2011sampling,
  title={Sampling-based algorithms for optimal motion planning},
  author={Karaman, Sertac and Frazzoli, Emilio},
  journal={The international journal of robotics research},
  volume={30},
  number={7},
  pages={846--894},
  year={2011},
  publisher={Sage Publications Sage UK: London, England}
}

@article{liu1992path,
  title={Path planning using a tangent graph for mobile robots among polygonal and curved obstacles: Communication},
  author={Liu, Yun-Hui and Arimoto, Suguru},
  journal={The International Journal of Robotics Research},
  volume={11},
  number={4},
  pages={376--382},
  year={1992},
  publisher={Sage Publications Sage CA: Thousand Oaks, CA}
}

@inproceedings{warren1993fast,
  title={Fast path planning using modified A* method},
  author={Warren, Charles W},
  booktitle={[1993] Proceedings IEEE International Conference on Robotics and Automation},
  pages={662--667},
  year={1993},
  organization={IEEE}
}

@article{jian2021global,
  title={A global-local coupling two-stage path planning method for mobile robots},
  author={Jian, Zhiqiang and Zhang, Songyi and Chen, Shitao and Nan, Zhixiong and Zheng, Nanning},
  journal={IEEE Robotics and Automation Letters},
  volume={6},
  number={3},
  pages={5349--5356},
  year={2021},
  publisher={IEEE}
}

@article{wang2020dynamics,
  title={Dynamics-constrained global-local hybrid path planning of an autonomous surface vehicle},
  author={Wang, Ning and Xu, Hongwei},
  journal={IEEE transactions on vehicular technology},
  volume={69},
  number={7},
  pages={6928--6942},
  year={2020},
  publisher={IEEE}
}

@article{han2024imaging,
  title={Imaging-guided bioresorbable acoustic hydrogel microrobots},
  author={Han, Hong and Ma, Xiaotian and Deng, Weiting and Zhang, Junhang and Tang, Songsong and Pak, On Shun and Zhu, Lailai and Criado-Hidalgo, Ernesto and Gong, Chen and Karshalev, Emil and others},
  journal={Science Robotics},
  volume={9},
  number={97},
  pages={eadp3593},
  year={2024},
  publisher={American Association for the Advancement of Science}
}

@article{xie2020photoacoustic,
  title={Photoacoustic imaging-trackable magnetic microswimmers for pathogenic bacterial infection treatment},
  author={Xie, Lisi and Pang, Xin and Yan, Xiaohui and Dai, Qixuan and Lin, Huirong and Ye, Jing and Cheng, Yi and Zhao, Qingliang and Ma, Xing and Zhang, Xianzhong and others},
  journal={Acs Nano},
  volume={14},
  number={3},
  pages={2880--2893},
  year={2020},
  publisher={ACS Publications}
}

@article{mnih2015human,
  title={Human-level control through deep reinforcement learning},
  author={Mnih, Volodymyr and Kavukcuoglu, Koray and Silver, David and Rusu, Andrei A and Veness, Joel and Bellemare, Marc G and Graves, Alex and Riedmiller, Martin and Fidjeland, Andreas K and Ostrovski, Georg and others},
  journal={nature},
  volume={518},
  number={7540},
  pages={529--533},
  year={2015},
  publisher={Nature Publishing Group UK London}
}
\bibliographystyle{sciencemag}

\paragraph*{Competing interests:}
There are no competing interests to declare.

\paragraph*{Code, data and materials availability:}
All data supporting the findings of this study are available in the paper and its Supplementary Materials. Code, raw microscopy videos and primary datasets are available from the corresponding author upon reasonable request.


\subsection*{Supplementary materials}
Supplementary Methods\\
Additional Validation\\
Figs. S1 to S8\\
Tables S1 to S4\\
Supplementary Videos 1--11


\newpage


\renewcommand{\thefigure}{S\arabic{figure}}
\renewcommand{\thetable}{S\arabic{table}}
\renewcommand{\theequation}{S\arabic{equation}}
\renewcommand{\thepage}{S\arabic{page}}
\setcounter{figure}{0}
\setcounter{table}{0}
\setcounter{equation}{0}
\setcounter{page}{1} 


\begin{center}
\section*{Supplementary Materials for\\ \scititle}

\end{center}

\subsubsection*{This PDF file includes:}
Supplementary Methods\\
Additional Validation\\
Figures S1 to S8\\
Tables S1 to S4

\subsubsection*{Other Supplementary Materials for this manuscript:}
Supplementary Videos 1--11

\newpage


\section*{Supplementary Methods}

\subsection*{Three-dimensional extension of AGP}

\begin{figure}[h]
    \centering
    \includegraphics[width=0.75\textwidth]{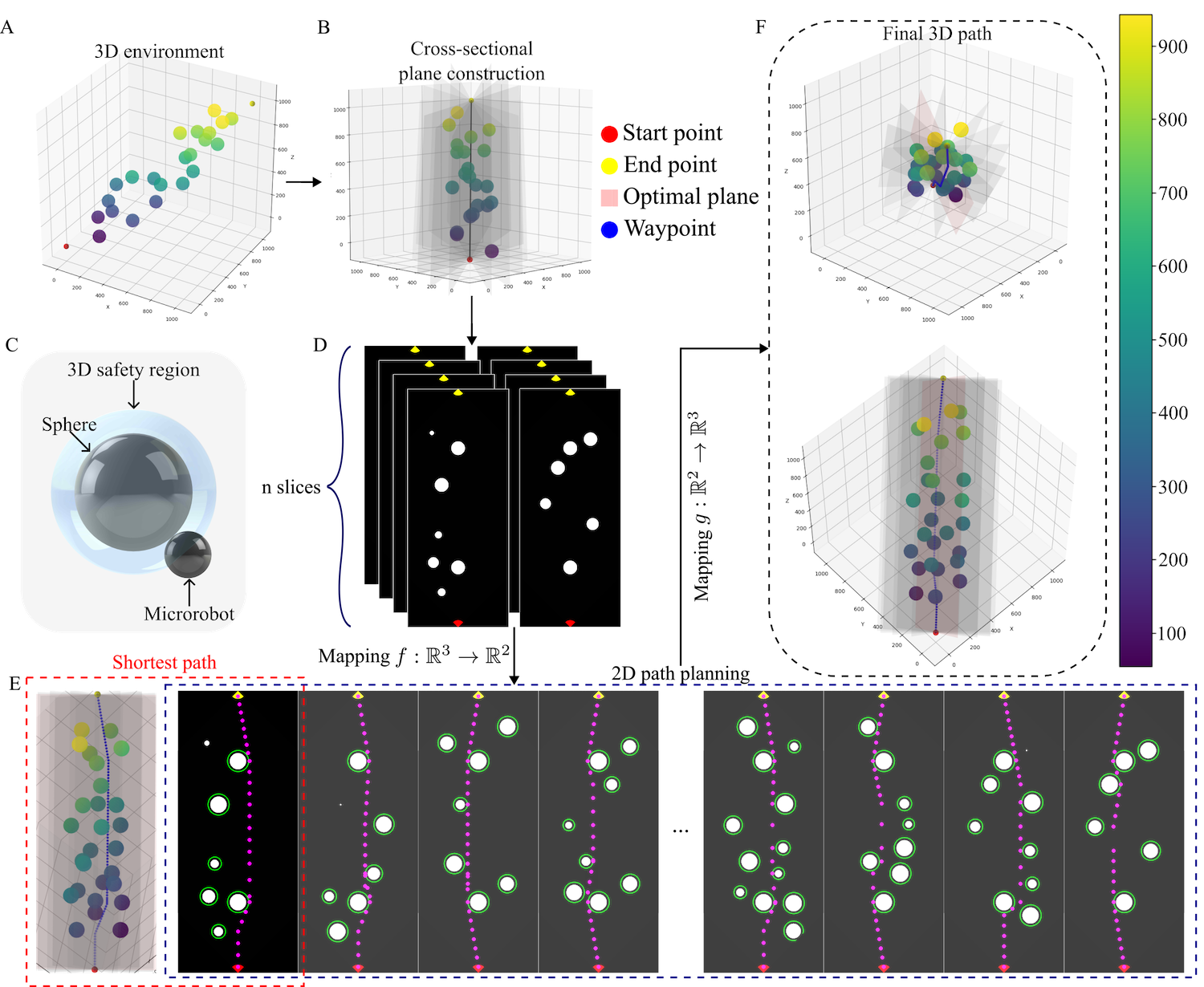}
    \caption{\textbf{Three-dimensional extension of AGP.} (\textbf{A,B}) A family of cross-sectional planes is generated about the start--goal direction and used to sample the three-dimensional workspace. (\textbf{C}) An inflated spherical obstacle intersects a candidate plane in a circular safety region. (\textbf{D}) Intersected obstacle geometry is mapped to two dimensions for planar AGP. (\textbf{E}) AGP generates a collision-free path on a candidate cross-section and the path length is evaluated across sampled planes. (\textbf{F}) The path from the selected plane is lifted back to the three-dimensional workspace.}
    \label{fig:3D}
\end{figure}

The three-dimensional extension evaluates a family of cross-sectional planes containing the start--goal direction, maps intersected obstacle geometry to each plane, applies the two-dimensional AGP on each slice, and lifts the selected planar solution back to three dimensions (Supplementary Fig.~\ref{fig:3D}).

Let
\[
\mathbf{S}=(x_s,y_s,z_s)^\top,\qquad
\mathbf{E}=(x_e,y_e,z_e)^\top,\qquad
\mathbf{d}=\mathbf{E}-\mathbf{S},\qquad
\hat{\mathbf{d}}=\frac{\mathbf{d}}{\|\mathbf{d}\|}.
\]
Choose a unit vector $\mathbf{q}_0$ orthogonal to $\hat{\mathbf{d}}$. For slice $i=1,\ldots,n$, define
\[
\phi_i=\frac{2\pi(i-1)}{n},
\]
rotate $\mathbf{q}_0$ about $\hat{\mathbf{d}}$ to obtain $\mathbf{q}_i$, and set
\[
\mathbf{e}_1=\hat{\mathbf{d}},\qquad
\mathbf{e}_2=\mathbf{q}_i,\qquad
\mathbf{n}_i=\mathbf{e}_1\times\mathbf{e}_2.
\]
The corresponding plane is
\[
\mathcal{P}_i=
\left\{
\mathbf{S}+u\mathbf{e}_1+v\mathbf{e}_2:\;u,v\in\mathbb{R}
\right\}.
\]

For spherical obstacle $m$ with center $\mathbf{C}_m$ and radius $R_m$, the inflated three-dimensional safety region is
\[
r^{3D}_m=R_m+R.
\]
Its signed distance from $\mathcal{P}_i$ is
\[
\delta_{i,m}=(\mathbf{C}_m-\mathbf{S})\cdot\mathbf{n}_i,
\qquad
d_{i,m}=|\delta_{i,m}|.
\]
If $d_{i,m}\leq r^{3D}_m$, the plane intersects the inflated sphere in a circle with center
\[
\mathbf{P}_{i,m}=\mathbf{C}_m-\delta_{i,m}\mathbf{n}_i
\]
and radius
\[
r^{\mathrm{int}}_{i,m}=
\sqrt{(r^{3D}_m)^2-d_{i,m}^2}.
\]

Using the plane basis, the projection and lifting maps are
\[
f_i(\mathbf{P})=
\begin{bmatrix}
\mathbf{e}_1^\top(\mathbf{P}-\mathbf{S})\\
\mathbf{e}_2^\top(\mathbf{P}-\mathbf{S})
\end{bmatrix},
\qquad
g_i(u,v)=\mathbf{S}+u\mathbf{e}_1+v\mathbf{e}_2.
\]
AGP is applied independently on each two-dimensional slice. If the waypoint sequence on slice $i$ is
$\{\mathbf{w}_{i,1},\ldots,\mathbf{w}_{i,k_i}\}$,
its path length is
\[
L_i=
\sum_{j=1}^{k_i-1}
\|\mathbf{w}_{i,j+1}-\mathbf{w}_{i,j}\|_2.
\]
The selected slice is
\[
i^\ast=\arg\min_{1\leq i\leq n}L_i,
\]
and the final three-dimensional path is reconstructed by applying $g_{i^\ast}$ to the selected planar waypoints. In the representative example, the full slicing, planar planning, slice selection, and reconstruction procedure required approximately $0.3\,\mathrm{s}$. This result is presented as a feasibility demonstration of the extension.

\subsection*{Image-processing workflow}

\begin{figure}[h]
    \centering
    \includegraphics[width=0.75\textwidth]{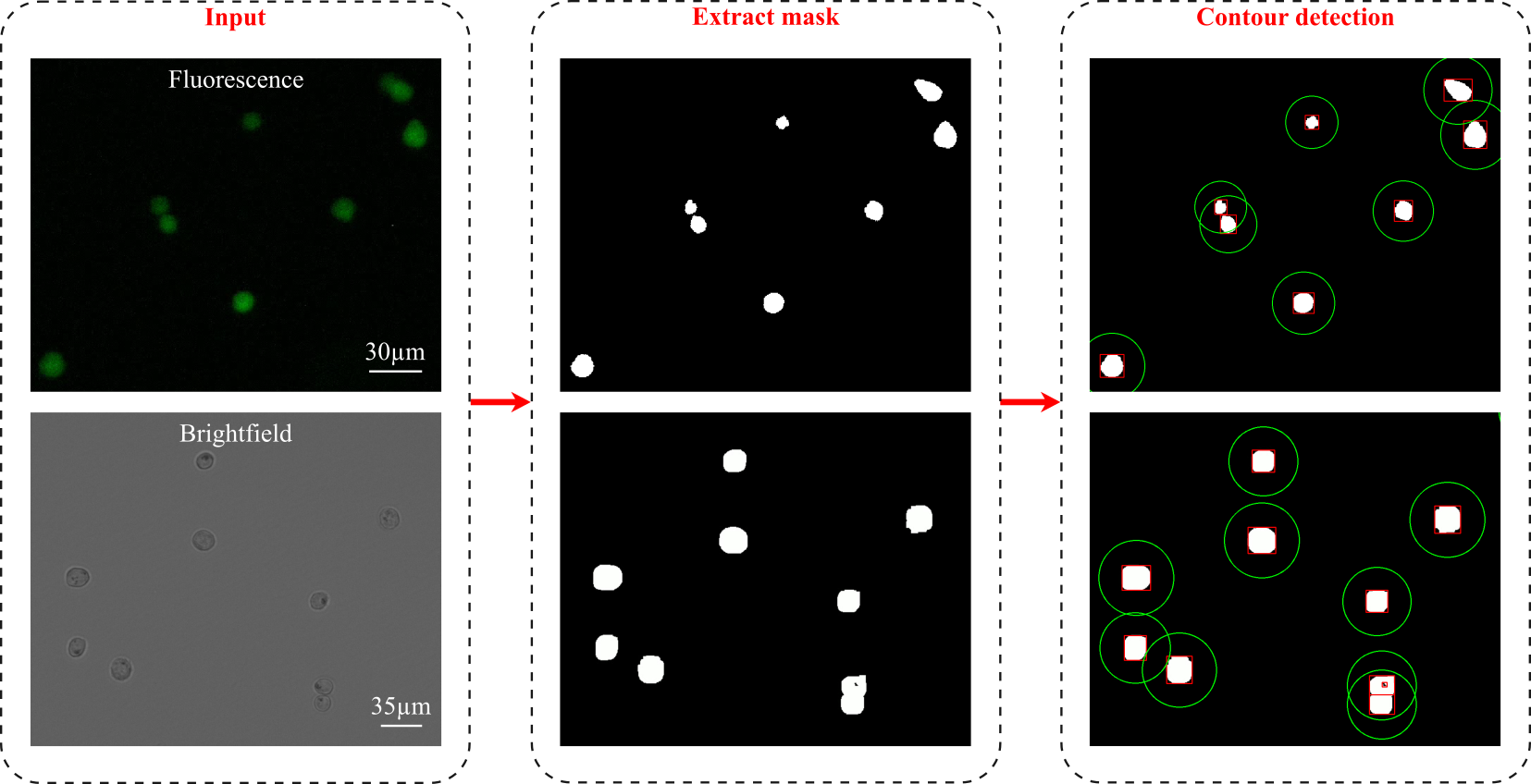}
    \caption{\textbf{Image-processing workflow for microrobot and obstacle detection.} Representative fluorescence and bright-field inputs are converted to binary masks by thresholding, followed by contour extraction and geometric state estimation. Green circles indicate the safety regions constructed from the detected object geometry for planning and local control.}
    \label{fig:safety}
\end{figure}

Each video frame was converted to grayscale for contour extraction; HSV representations were additionally used when color-based discrimination was required. Thresholding generated binary masks for the microrobot, obstacles, and relevant workspace boundaries. Contours were extracted from these masks, and axis-aligned bounding boxes were used to obtain object centers and dimensions. The resulting geometric states were passed to the planner and controller. Supplementary Fig.~\ref{fig:safety} shows representative fluorescence and bright-field inputs, the corresponding binary masks, and contour detection with safety-region overlays.

\subsection*{Baseline planner implementations}

For the static benchmark, all planners received the same rasterized obstacle mask, start--goal pair, moving-body radius, and optional padding through the shared planning interface.

\subsubsection*{Weighted A*}
Weighted A* operated on an 8-connected pixel grid. Obstacle contours were extracted from the common raster mask and represented by minimum-enclosing circles; each circle was inflated by the moving-body radius and any additional padding before rasterization into the occupancy mask \cite{pohl1970heuristic}. The evaluation function was
\begin{equation}
    f(n)=g(n)+w\,h(n),
\end{equation}
with heuristic weight $w=1.1$ and Euclidean heuristic
\begin{equation}
    h(n)=\sqrt{(x_e-x_n)^2+(y_e-y_n)^2}.
\end{equation}
Horizontal and vertical moves had unit cost and diagonal moves had cost $\sqrt{2}$. If rasterization placed the start pixel inside an inflated obstacle, the implementation searched for the nearest free pixel within an 80-pixel radius; a goal inside an inflated obstacle was treated as infeasible.

\subsubsection*{RRT*}
RRT* used the same inflated occupancy mask for collision checking \cite{karaman2011sampling}. Each tree extension was limited to 30 pixels, the goal tolerance was 30 pixels, and the planner used a maximum of 3,000 sampling iterations. Rewiring used a fixed 80-pixel neighborhood, and the goal was sampled directly with probability 0.08. Candidate edges were checked by dense sampling along the line segment against the occupancy mask. The planner continued through the full iteration budget while retaining the lowest-cost goal connection found. The resulting path was post-processed by 120 random shortcut attempts.

\subsubsection*{Circular-obstacle pathfinding}
The COP implementation first converted the common raster mask to inflated circular obstacles using the same minimum-enclosing-circle procedure used by the other baseline wrappers. Start and goal were appended as zero-radius circles. A geometry graph was then constructed from valid internal and external bitangents between circle pairs; a bitangent edge was retained only when the segment had line of sight with respect to all other circles \cite{liu1992path}. Nodes lying on the same obstacle circle were additionally connected by ``hugging'' edges. Graph search used zero heuristic, that is, Dijkstra search, with edge costs
\begin{equation}
    c(a,b)=
    \begin{cases}
        1+\|\mathbf{a}-\mathbf{b}\|_2, & \text{for straight bitangent edges},\\
        1+r\,\Delta\theta, & \text{for same-circle arc edges},
    \end{cases}
\end{equation}
where $r\,\Delta\theta$ is the shorter-arc length. The additive unit term follows the implemented reference-style graph cost and mildly favors solutions containing fewer graph edges. Same-circle portions of the final path were expanded along the shorter circular arc with approximately 4-pixel sampling.

\begin{table}[h]
\centering
\caption{\textbf{Simulation environment parameters used for DQN training.}}
\label{tab:environment}
\begin{tabular}{@{}ll@{}}
\hline
\textbf{Parameter (symbol)} & \textbf{Value} \\
\hline
Arena dimensions ($W\times H$) & $2000\,\mathrm{px}\times2000\,\mathrm{px}$ \\
Microrobot radius ($R$) & $25\,\mathrm{px}$ \\
Microrobot displacement per step ($v_m$) & $10\,\mathrm{px/step}$ \\
Dynamic obstacles ($N_d$) & 50 \\
Static obstacles ($N_s$) & 5 \\
Obstacle radius ($r_i$) & $25$--$50\,\mathrm{px}$ \\
Dynamic-obstacle displacement per step ($v_i$) & $4$--$8\,\mathrm{px/step}$ \\
Simulation time step ($\Delta$) & $0.04\,\mathrm{s}$ \\
\hline
\end{tabular}
\end{table}

\begin{table}[h]
\centering
\caption{\textbf{Reward-function parameters used for DQN training.}}
\label{tab:function}
\begin{tabular}{@{}ll@{}}
\hline
\textbf{Parameter (symbol)} & \textbf{Value} \\
\hline
Shaping reward weight ($k_{\mathrm{shape}}$) & 1.0 \\
Time penalty ($\lambda_{\mathrm{step}}$) & 0.05 \\
Collision penalty ($R_{\mathrm{collision}}$) & $-50$ \\
Success reward ($R_{\mathrm{success}}$) & 50 \\
\hline
\end{tabular}
\end{table}

\begin{table}[h]
\centering
\caption{\textbf{DQN training hyperparameters.}}
\label{tab:DQN}
\begin{tabular}{@{}ll@{}}
\hline
\textbf{Parameter (symbol)} & \textbf{Value} \\
\hline
Number of parallel environments ($n_{\mathrm{env}}$) & 8 \\
Replay buffer capacity ($N$) & 500,000 \\
Batch size ($B$) & 256 \\
Initial exploration steps ($S_{\mathrm{ini}}$) & 5,000 \\
Network update frequency ($f$) & 4 \\
Target-network update interval ($K$) & 1,000 \\
Discount factor ($\gamma$) & 0.99 \\
Initial exploration rate ($\varepsilon_0$) & 1 \\
Final exploration rate ($\varepsilon_1$) & 0.01 \\
$\varepsilon$-annealing fraction ($\alpha_\varepsilon$) & 40\% \\
Learning rate ($\lambda_0$) & $1\times10^{-4}\rightarrow0$ \\
Hidden-layer sizes & [512, 256] \\
Evaluation frequency ($F$) & 5,000 \\
Evaluation episodes ($M$) & 50 \\
Total training steps ($S$) & $1\times10^6$ \\
\hline
\end{tabular}
\end{table}

\subsection*{Magnetic actuation and waypoint tracking}

A rotating magnetic field produces a torque
\begin{equation}
    \boldsymbol{\tau}(t)=\boldsymbol{m}\times\boldsymbol{B}(t),
\end{equation}
where $\boldsymbol{m}$ is the microrobot magnetic moment. The three-axis field was generated as
\begin{equation}
\label{eq:B_original}
\boldsymbol{B}(t)=
B_0
\begin{bmatrix}
-\cos\beta\,\cos\alpha\,\cos\omega t+\sin\alpha\,\sin\omega t\\
-\cos\beta\,\sin\alpha\,\cos\omega t-\cos\alpha\,\sin\omega t\\
\sin\beta\,\cos\omega t
\end{bmatrix},
\end{equation}
where $B_0$ is the field amplitude, $\omega$ is the angular frequency, $\alpha$ is the azimuth of the rotation axis in the $xy$-plane, and $\beta$ is the polar angle of that axis from $+z$. The field magnitude satisfies $\|\boldsymbol{B}(t)\|=B_0$. For planar rolling, $\beta=\pi/2$ was used so that the rotation axis lay in the substrate plane.

\begin{figure}[h]
    \centering
    \includegraphics[width=0.75\textwidth]{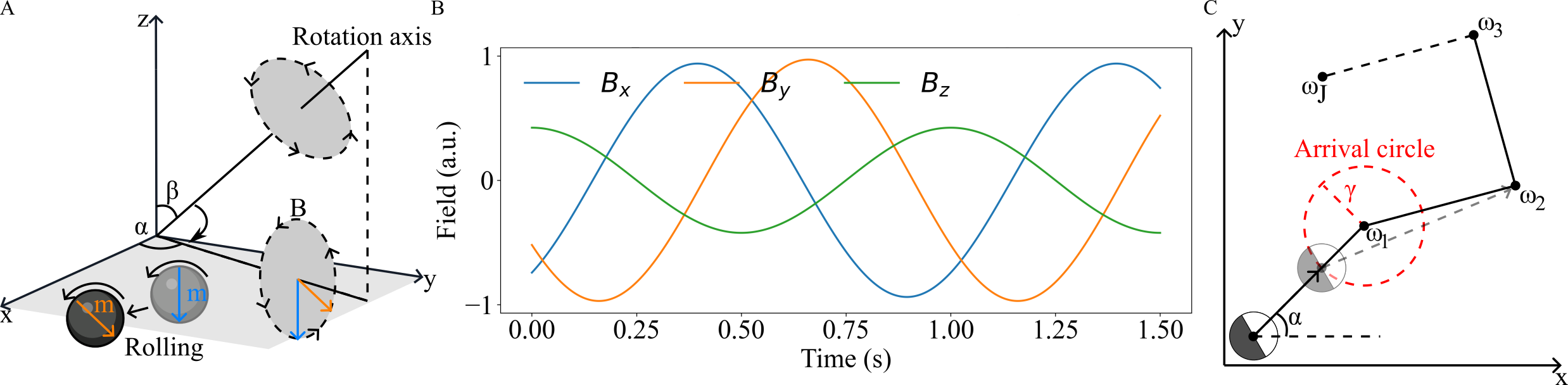}
    \caption{\textbf{Magnetic actuation and closed-loop waypoint tracking.} (\textbf{A}) Geometry of the rotating magnetic field. The rotation axis is parameterized by azimuth $\alpha$ and polar angle $\beta$, and $\boldsymbol{B}(t)$ rotates in the plane normal to this axis. (\textbf{B}) Representative normalized field components for $\alpha=35^\circ$ and $\beta=25^\circ$. (\textbf{C}) Planar waypoint tracking: an arrival region of radius $\gamma$ determines the switch from waypoint $\boldsymbol{w}_j$ to $\boldsymbol{w}_{j+1}$. Planar rolling experiments used $\beta=\pi/2$.}
    \label{fig:control}
\end{figure}

Let the current microrobot position be $\boldsymbol{m}=(x_m,y_m)^\top$ and the active waypoint be $\boldsymbol{w}_j=(x_{w,j},y_{w,j})^\top$. The waypoint bearing and distance were
\begin{equation}
    \theta_j=\operatorname{atan2}(y_{w,j}-y_m,x_{w,j}-x_m),
    \qquad
    d_j=\|\boldsymbol{w}_j-\boldsymbol{m}\|_2.
\end{equation}
When $d_j\leq\gamma$, the controller advanced to the next waypoint. Otherwise, the in-plane rotation axis was selected perpendicular to the desired rolling direction according to the system sign convention, and the rotating field was updated accordingly. Supplementary Fig.~\ref{fig:control} summarizes the field geometry and waypoint-switching rule.

\begin{table}[h]
\centering
\caption{\textbf{Helmholtz-coil geometry.}}
\label{tab:helmholtz}
\begin{tabular}{@{}ll@{}}
\hline
\textbf{Parameter} & \textbf{Value} \\
\hline
Wire gauge & 24 AWG copper \\
Turns per coil $(N_s,N_m,N_l)$ & $\approx(368,368,260)$ \\
Small-pair separation / radius & $26\,\mathrm{mm}$ / $26\,\mathrm{mm}$ \\
Medium-pair separation / radius & $66\,\mathrm{mm}$ / $35\,\mathrm{mm}$ \\
Large-pair separation / radius & $84\,\mathrm{mm}$ / $54\,\mathrm{mm}$ \\
Mounting orientation & small: horizontal; medium/large: vertical \\
\hline
\end{tabular}
\end{table}

\subsection*{Computational environment}

Simulations were run on a MacBook equipped with an Apple M3 Pro processor (12 CPU cores, 18 GB unified memory). Experiments were executed on a 64-bit Windows workstation with an Intel Core i7-9700K processor at 3.60 GHz and 16 GB RAM. Algorithms were implemented in Python~3.10. OpenCV was used for image processing and visualization, and the reinforcement-learning implementation used Gymnasium and Stable-Baselines3.

\section*{Additional Validation}

\begin{figure}[h]
    \centering
    \includegraphics[width=0.75\textwidth]{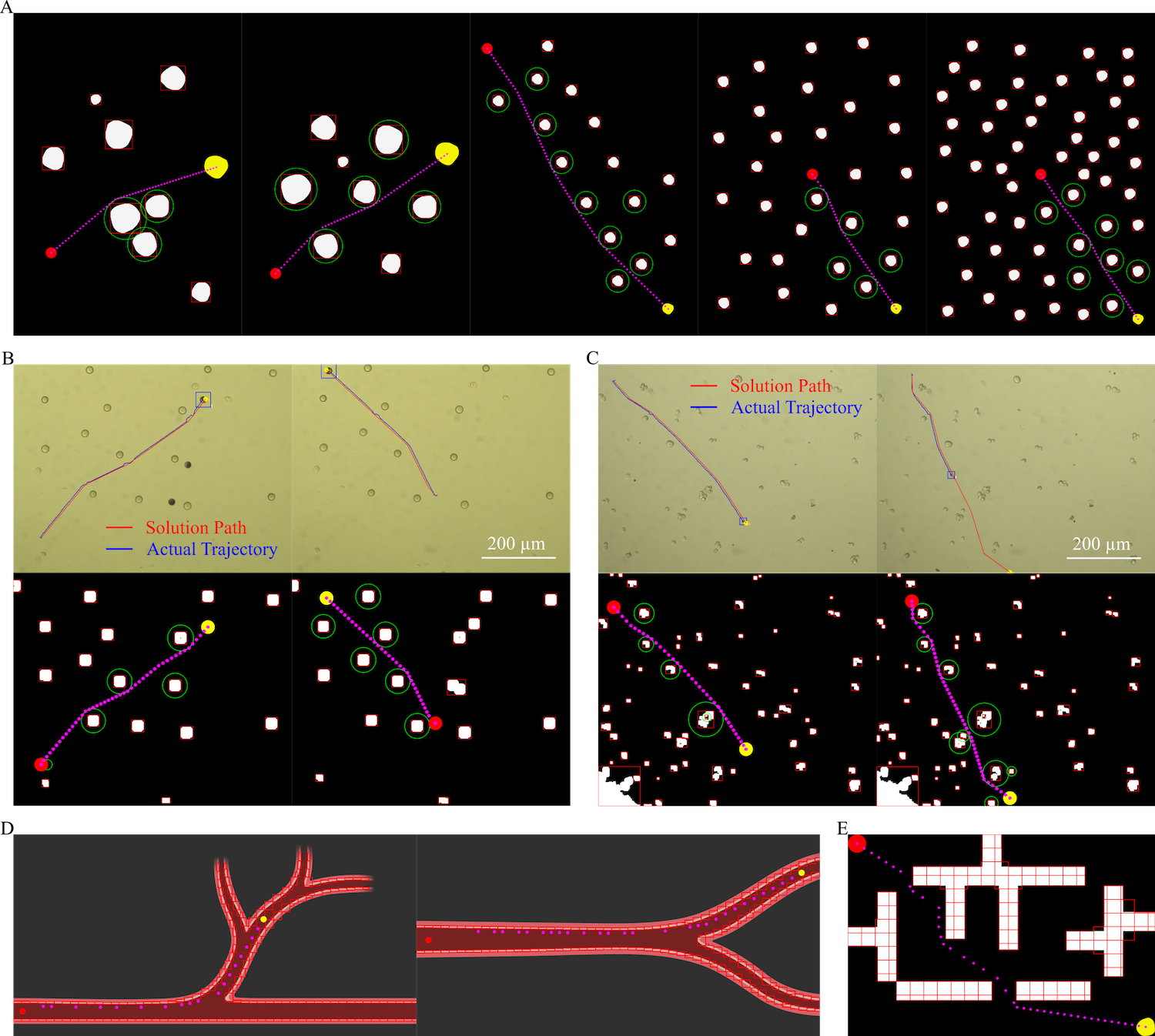}
    \caption{\textbf{Additional validation of AGP in static environments.} (\textbf{A}) Simulated environments with increasing obstacle density. (\textbf{B}) Navigation experiments in a silica-particle field. Top, bright-field images overlaid with the AGP solution path (red) and measured microrobot trajectory (blue); bottom, the corresponding binary masks used for planning. (\textbf{C}) Navigation experiments in a CHO-cell field, with the corresponding bright-field images and planning masks. (\textbf{D}) Simulated vascular geometries in which vessel walls are represented as boundary obstacles. (\textbf{E}) Maze-like environment with a valid AGP route from start to target.}
    \label{fig:AGP_static_additional}
\end{figure}

\begin{figure}[h]
    \centering
    \includegraphics[width=0.75\textwidth]{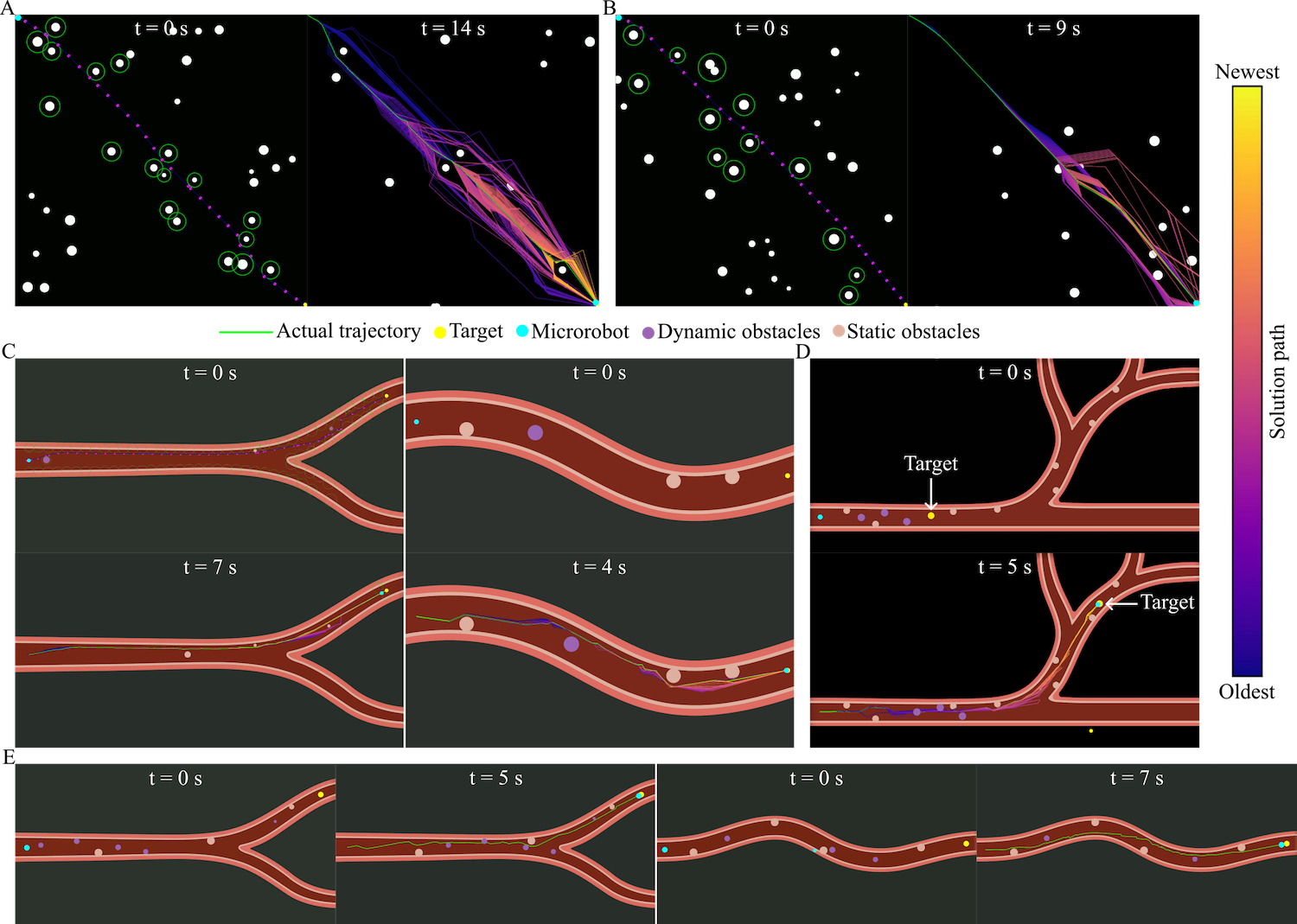}
    \caption{\textbf{Successful online AGP replanning without a local escape controller in dynamic environments.} (\textbf{A}) Randomly moving obstacles. (\textbf{B}) Obstacles translating in a common direction. (\textbf{C}) Two vascular geometries containing dynamic obstacles. (\textbf{D}) Vascular geometry with a moving target. (\textbf{E}) Two additional vascular geometries containing dynamic obstacles. In these examples, AGP is recomputed online as the obstacle or target configuration changes, allowing global replanning alone to adapt the solution path without activation of a local escape controller.}
    \label{fig:AGP_online_dynamic}
\end{figure}

\begin{figure}[h]
    \centering
    \includegraphics[width=0.75\textwidth]{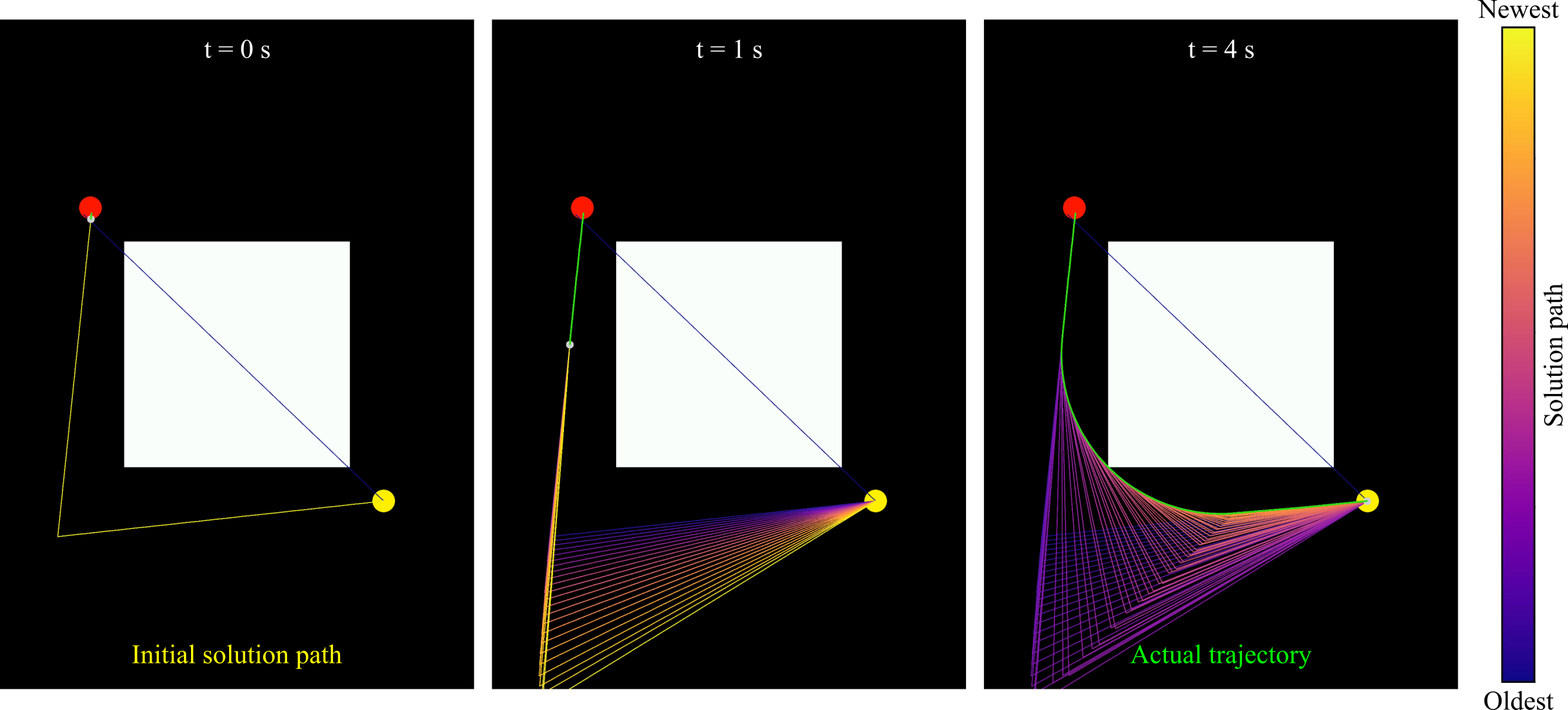}
    \caption{\textbf{Online AGP replanning can shorten the executed route in a static environment.} The initial solution path at $t=0$ is repeatedly recomputed from the updated microrobot position as navigation proceeds ($t=1$ s and $t=4$ s). Successive solution paths are colored from oldest (purple) to newest (yellow), and the measured microrobot trajectory is shown in green. Because each new plan is constructed from the current microrobot position rather than following the initial path as a fixed route, online replanning progressively updates the remaining path and yields an executed trajectory shorter than the original $t=0$ solution.}
    \label{fig:AGP_online_static}
\end{figure}

\begin{figure}[h]
    \centering
    \includegraphics[width=0.75\textwidth]{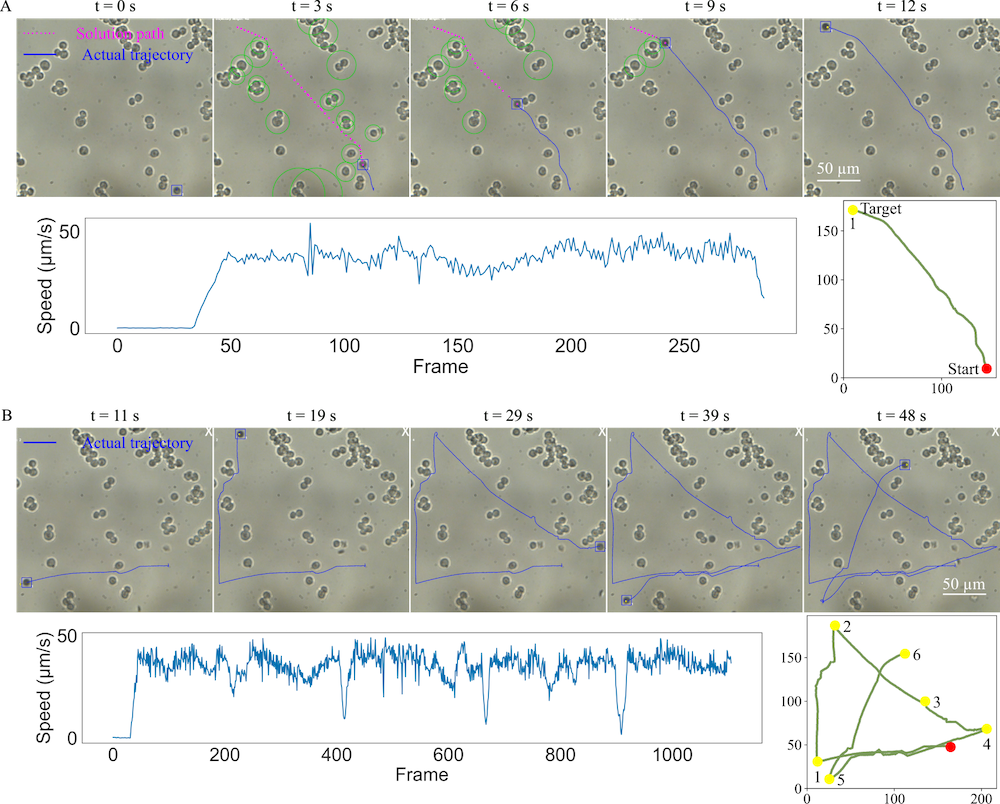}
    \caption{\textbf{Additional experimental validation of online AGP replanning in a static CHO-cell environment.} Time-lapse bright-field images show the planned path and measured microrobot trajectory, together with the corresponding microrobot-speed and planar-trajectory histories. (\textbf{A}) Single-target navigation. (\textbf{B}) Sequential navigation among multiple targets. In both cases, AGP is repeatedly recomputed from the measured microrobot position during closed-loop navigation.}
    \label{fig:AGP_online_experiment}
\end{figure}

\begin{figure}[h]
    \centering
    \includegraphics[width=0.75\textwidth]{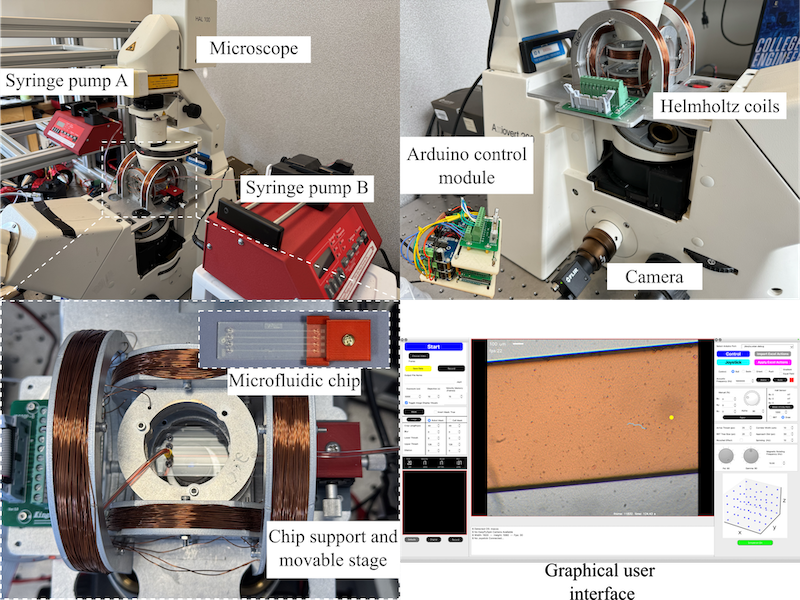}
    \caption{\textbf{Experimental platform for autonomous magnetic microrobot navigation.} The benchtop system integrates syringe pumps for controlled flow, an inverted microscope and camera for real-time imaging, a Helmholtz-coil assembly for magnetic actuation, an Arduino-based control module, and a microfluidic chip mounted on a movable stage within the actuation workspace. The graphical user interface integrates image acquisition, navigation, and magnetic control during closed-loop experiments.}
    \label{fig:experimental_platform}
\end{figure}



\end{document}